\documentclass{article}

\usepackage[main, final]{neurips_2026}

\usepackage[utf8]{inputenc}
\usepackage[T1]{fontenc}
\usepackage{microtype}
\usepackage{graphicx}
\usepackage{subcaption}
\usepackage{booktabs}
\usepackage{hyperref}
\usepackage{url}
\usepackage{amsmath}
\usepackage{amssymb}
\usepackage{mathtools}
\usepackage{amsthm}
\usepackage{xcolor}
\usepackage{nicefrac}
\usepackage{smile}
\usepackage[capitalize,noabbrev]{cleveref}
\usepackage{wrapfig}

\usepackage{titlesec}
\AtBeginDocument{%
  \addtolength\abovedisplayskip{-0.25\baselineskip}%
  \addtolength\belowdisplayskip{-0.25\baselineskip}%
  \addtolength\abovedisplayshortskip{-0.25\baselineskip}%
  \addtolength\belowdisplayshortskip{-0.25\baselineskip}%
}
\titlespacing*{\section}{0pt}{0.0\baselineskip}{0.0\baselineskip}
\titlespacing*{\subsection}{0pt}{0.0\baselineskip}{0.0\baselineskip}
\newcommand{\INPUT}[1]{\REQUIRE #1}
\newcommand{\OUTPUT}[1]{\ENSURE #1}
\newcommand{\our}{IDEA Prune}

\title{On the Efficiency of Structured Pruning in Small Language Model Pretraining}

\author{%
  Yixiao Li\textsuperscript{1}\thanks{Work done at Apple.} \quad
  Xianzhi Du\textsuperscript{2} \quad
  Ajay Kumar Jaiswal\textsuperscript{2} \quad
  Tao Lei\textsuperscript{2} \\
  \textbf{Tuo Zhao\textsuperscript{1} \quad
  Chong Wang\textsuperscript{2}\footnotemark[1] \quad
  Jianyu Wang\textsuperscript{2}\footnotemark[1]} \\
  \textsuperscript{1}Georgia Tech \quad \textsuperscript{2}Apple Inc. \\
  \texttt{yixiaoli@gatech.edu}
}

\begin{document}

\maketitle

\begin{abstract}
Companies that deploy generative language models on edge devices often need to train small models from scratch due to legal, data-governance, or product-customization requirements.
An alternative approach is to train an enlarged model and compress it to the target size via structured pruning, but this advantage becomes unclear once the cost of pretraining the enlarged model is included.
We introduce the enlarge-and-prune framework to study this question under full-cost accounting, and show that conventional enlarge-and-prune pipelines---which train the enlarged model, prune it, and recover in separate stages---do not consistently outperform direct small-model training under the same total token budget.
To overcome this limitation, we propose IDEA Prune, an integrated enlarge-and-prune pipeline that unifies enlarged model pretraining, iterative structured pruning, and recovery under a single cosine annealing learning rate schedule.
In controlled experiments compressing 2.8B models to 1.3B with up to 2T tokens on the same dataset, IDEA Prune consistently outperforms conventional pipelines and modestly improves overall performance relative to the scratch baseline, improving MMLU accuracy to 46.4\%, compared with 31.4--33.4\% for conventional pipelines.
These advantages are preserved after downstream supervised fine-tuning, where IDEA Prune matches or outperforms all baselines of the same size on instruction following, math reasoning, and code generation.
Our analysis further reveals that intermediate checkpoints provide more favorable pruning starting points than fully converged ones, and that an enlargement factor of approximately 2.6$\times$ yields the best pruned models in the tested regime.
\end{abstract}

\section{Introduction}
\label{sec:intro}

The deployment of generative language models on resource-constrained devices has become increasingly critical for modern applications.
Tech companies face growing demands to deploy generative language models on smartphones and edge devices \citep{abdin2024phi, gunter2024apple}.
Given the severe memory limitations of these devices, model sizes must remain small, typically under 3 billion parameters for practical deployment \citep{liu2024mobilellm, team2024gemini, lozhkov2025smollm2, deng2025plm}.

To obtain small models, prior work suggests two broad strategies.
One is to train the target small model directly with more data or more training tokens \citep{kaplan2020scaling, hoffmann2022training}.
The other is to start from a larger pretrained model and compress it to the target size.
Structured pruning is a common approach in the second strategy: it removes selected model structures, so that the compressed model matches the desired deployment architecture \citep{wang2019structured, xia2022structured, kwon2022fast}.
When the larger model is already available, pruning can be attractive because it may transfer the benefit of a higher-capacity source model without increasing the inference cost of the final small model \citep{xia2023sheared, muralidharan2024compact}.

In this paper, however, we do not assume access to existing large checkpoints in order to match a more realistic industrial production setting.
First, if the checkpoint is external, it may not be a valid input to an in-house training pipeline because of legal compliance, data governance, or product customization.
Indeed, recent industrial on-device and small-model efforts often train their models within company-controlled pipelines rather than relying on arbitrary external checkpoints \citep{gunter2024apple, team2024gemini, liu2024mobilellm}.
Second, if the checkpoint is internal, it may still be unsuitable as a pruning source.
For example, internal cloud-serving models are often orders of magnitude larger than on-device targets: cloud models can exceed 100B parameters, whereas practical on-device models must remain below a few billion parameters \citep{team2024gemini, liu2024mobilellm, lozhkov2025smollm2, deng2025plm}.
This size gap is not only a bookkeeping issue: in our experiments in Section~\ref{sec:model_size}, even moderate enlargement factors beyond $5\times$ lead to clear pruning degradation, suggesting that compression ratios of tens of times would be even less practical.
Thus, an internal checkpoint is not necessarily a suitable pruning source simply because it exists.

Given the aforementioned reasons, using pruning requires a different setup in the production setting: the source model must be deliberately trained as a moderately enlarged intermediate model and then pruned back to the deployment architecture.
Because this intermediate model is introduced only to produce the final small model and is never deployed, its training cost must be counted in the end-to-end production budget.
However, once this cost is included, is pruning still better than simply training the target small model from scratch?
This paper studies this question under the same total token budget and training data.

\begin{figure}[htb!]
\centering
\vspace{-0.175in}
\begin{subfigure}{0.42\linewidth}
    \centering
    \includegraphics[width=1.0\linewidth]{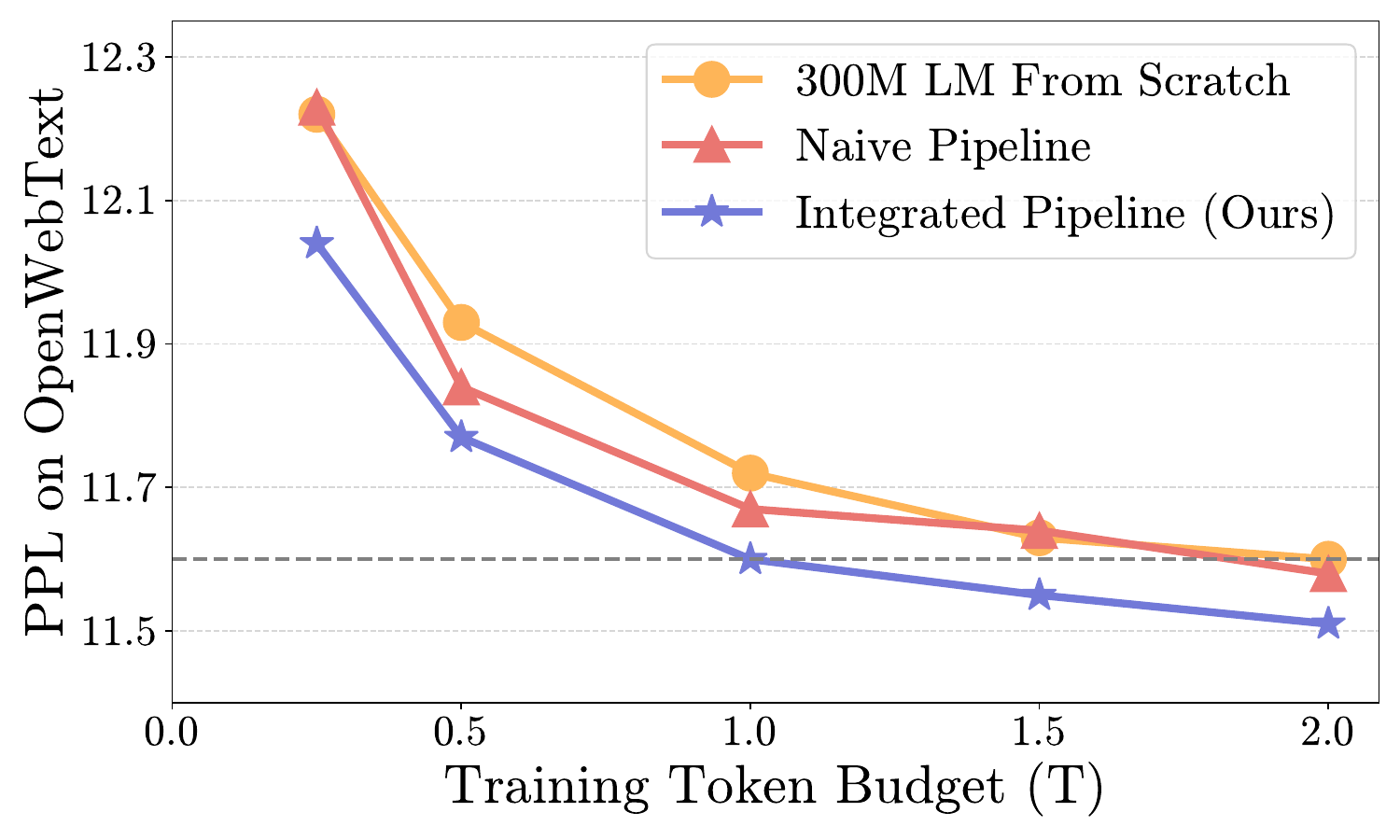}
    \caption{\label{fig:loss_curve_multi_run} Perplexity on OpenWebText}
\end{subfigure}
\begin{subfigure}{0.42\linewidth}
    \centering
    \includegraphics[width=1.0\linewidth]{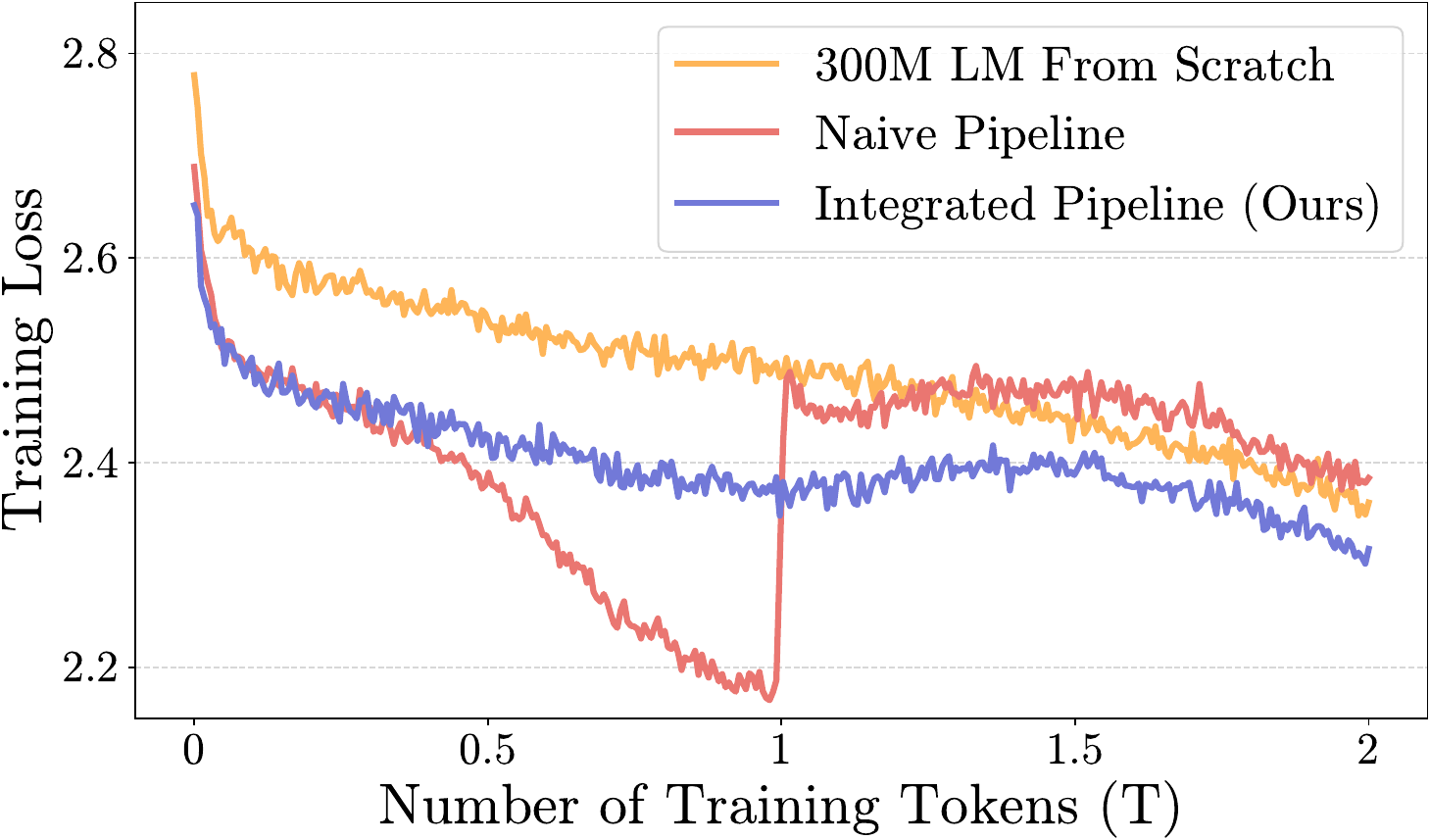}
    \caption{\label{fig:loss_curve_single_run} Training Loss}
\end{subfigure}
\vspace{-2mm}
\caption{\label{fig:loss_curve}
{\it Naive pipeline}: 1T for enlarged model training and 1T for pruned model recovery, each with separate learning rate decay.
{\it Integrated pipeline}: single learning rate schedule matching target-size model training, pruning started at 1T.
{\bf Left}: Perplexity on OpenWebText of 300M model trained with multiple training token budgets from 0.25T to 2T. The integrated pipeline is {\bf 2x} as token efficient as the naive pipeline.
{\bf Right}: Training loss curves for enlarge-and-prune pipelines. Training 300M models with 2T total token budget.}
\vspace{-0.2in}
\end{figure}

\begin{figure}[htb!]
\vspace{-0.15in}
\centering
\begin{subfigure}{0.3\linewidth}
    \centering
    \includegraphics[width=1.0\linewidth]{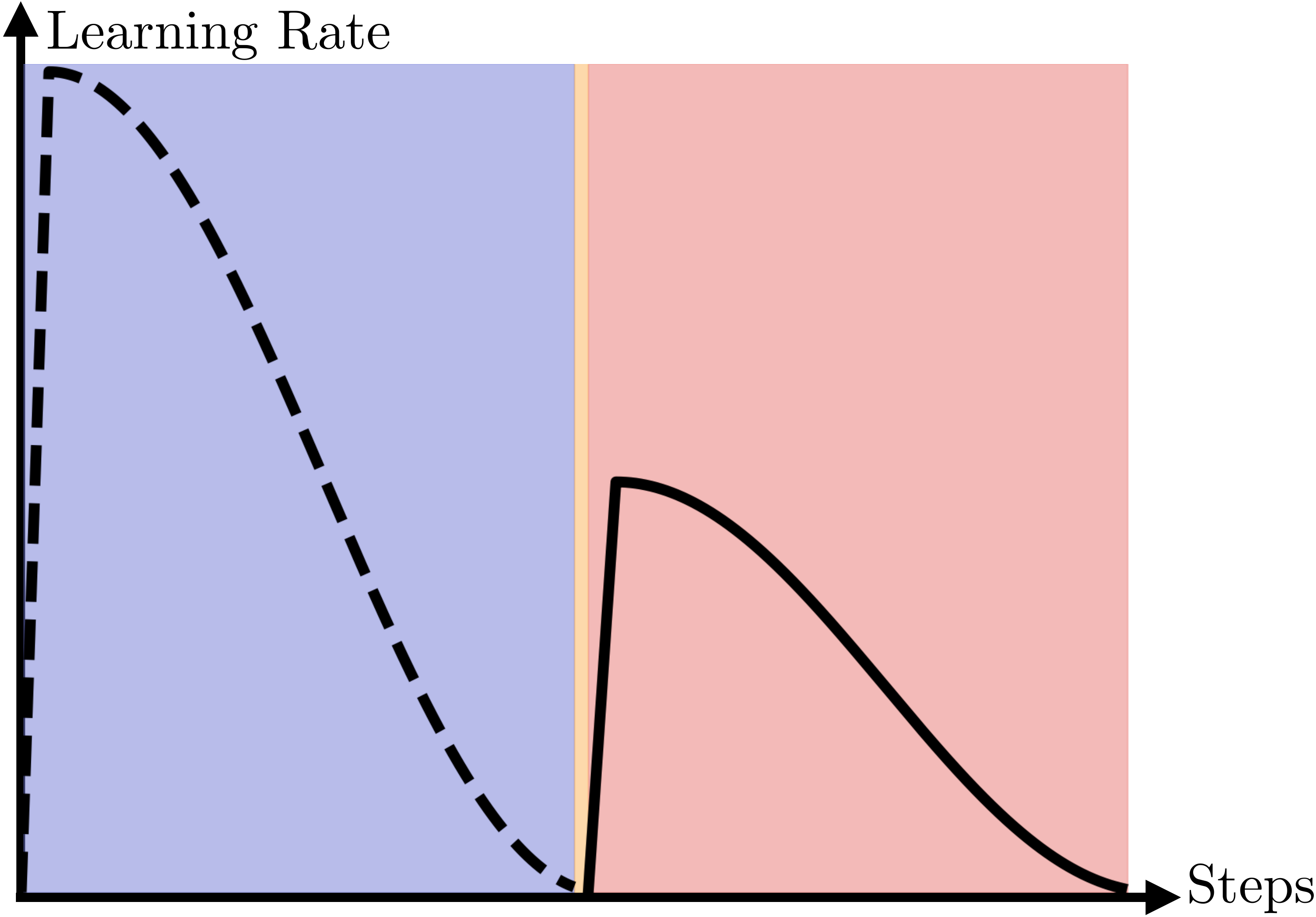}
    \caption{\label{fig:lr_minitron}Minitron}
\end{subfigure}
\hfill
\begin{subfigure}{0.3\linewidth}
    \centering
    \includegraphics[width=1.0\linewidth]{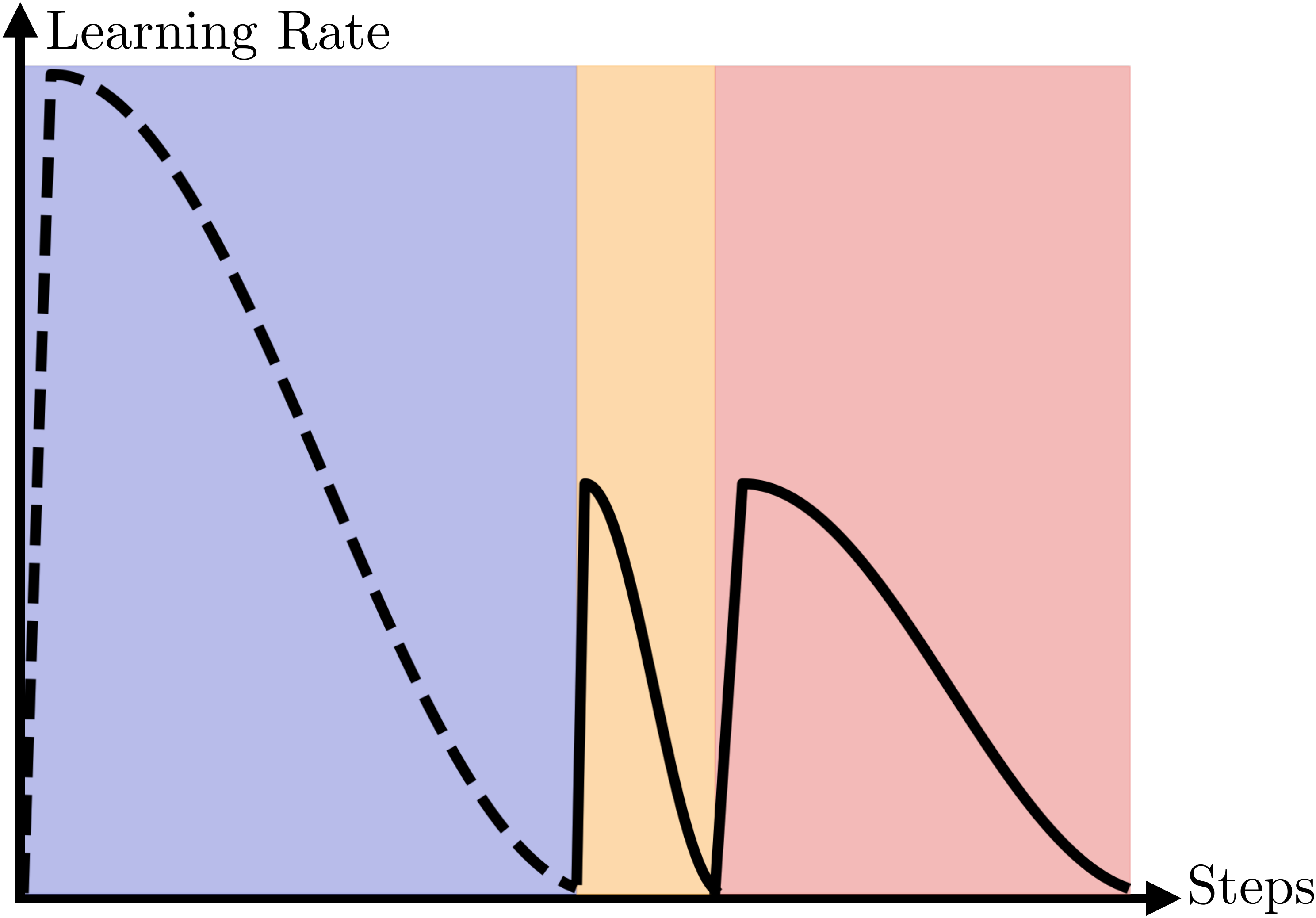}
    \caption{\label{fig:lr_sheared} Sheared LLaMA}
\end{subfigure}
\hfill
\begin{subfigure}{0.36\linewidth}
    \centering
    \includegraphics[width=1.0\linewidth]{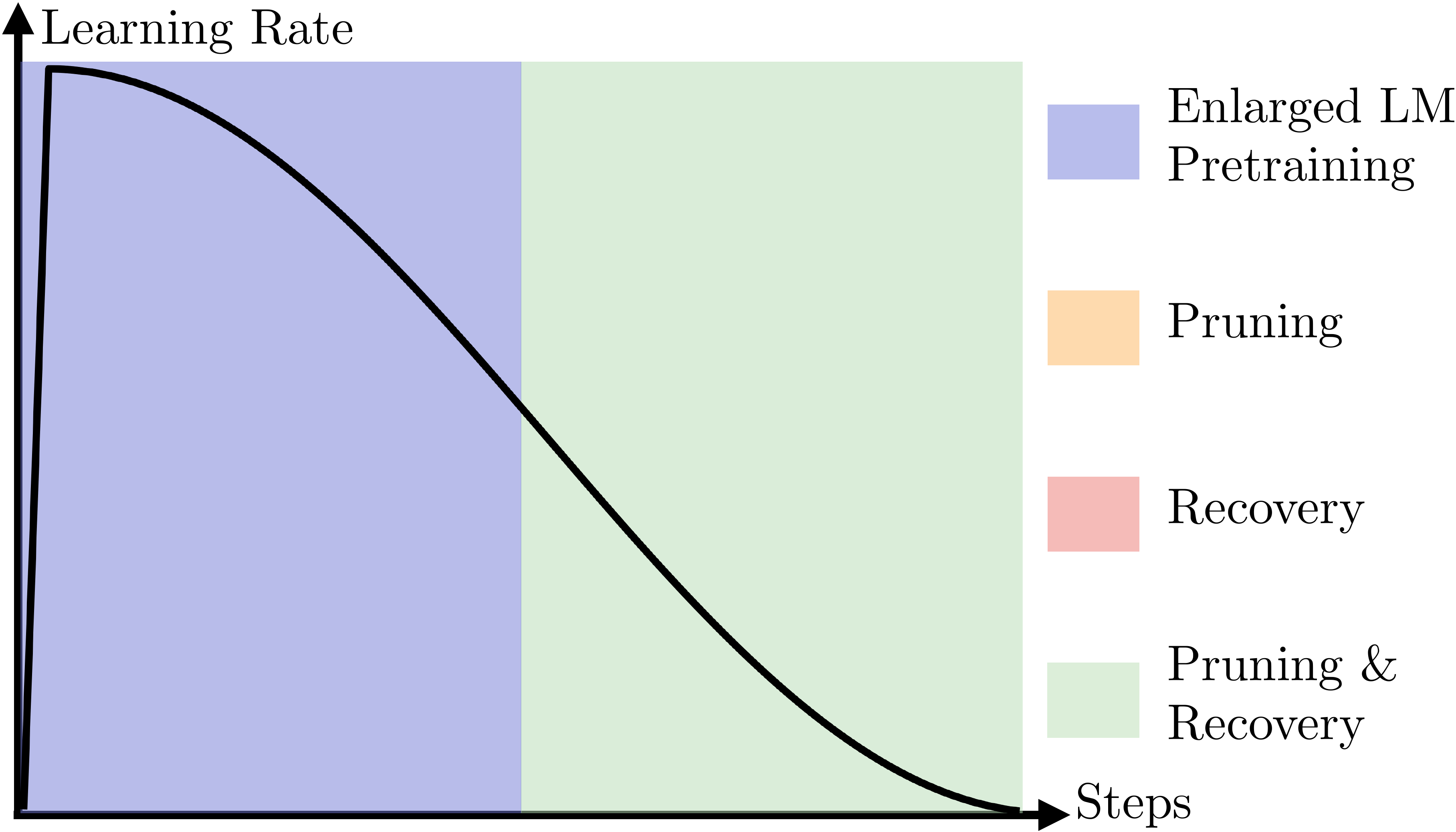}
    \caption{\label{fig:lr_integrated} IDEA Prune}
\end{subfigure}
\vspace{-0.1in}
\caption{\label{fig:main_fig} Pipeline overview and learning rate schedules. Naive pipelines (Figure~\ref{fig:lr_minitron}, \ref{fig:lr_sheared}) use separate LR schedules per stage, causing restarts and knowledge loss. {\our} (Figure~\ref{fig:lr_integrated}) unifies all stages---enlarged pretraining, iterative pruning, and recovery---under a single cosine annealing schedule, gradually reducing model size without LR restarts.}
\vspace{-0.1in}
\end{figure}

To enable a rigorous comparison, we introduce the {\it enlarge-and-prune framework} (Section~\ref{sec:separate_pruning_pipeline}), which accounts for the full cost of training the enlarged model.
Under this framework, we first examine a {\it naive enlarge-and-prune pipeline} that directly composes enlarged model pretraining with conventional two-stage pruning, and compare it against training the same small model from scratch under a matched token budget and on the same dataset.
Figure~\ref{fig:loss_curve_multi_run} shows that the naive pipeline fails to consistently achieve superior token efficiency compared to direct pretraining of small models.
This inefficiency stems from the divided nature of the naive pipeline: each stage uses its own learning rate schedule (Figure~\ref{fig:lr_minitron}, \ref{fig:lr_sheared}), causing sharp loss increases at stage boundaries (Figure~\ref{fig:loss_curve_single_run}) and catastrophic forgetting.
Furthermore, naive pipelines often employ disparate training objectives between stages, such as auxiliary regularization losses for mask learning \citep{xia2023sheared} and activation-based proxy losses \citep{muralidharan2024compact}, leading to suboptimal initialization for the recovery stage.

To overcome these limitations, we propose IDEA Prune, an {\bf I}ntegrate{\bf D} {\bf E}nlarge-{\bf A}nd-{\bf Prune} pipeline with a novel iterative structured pruning for producing small language models from scratch.
The integrated pipeline combines the enlarged model pretraining, pruning, and pruned model recovery stages under one cosine annealing learning rate schedule (Figure~\ref{fig:lr_integrated}), mitigating the knowledge loss caused by learning rate restarts in naive pipelines \citep{muralidharan2024compact, xia2023sheared}.
Furthermore, we extend iterative pruning \citep{zhu2017prune,louizos2017learning} to structured Feed-Forward Network (FFN) width compression.
Our approach progressively removes neurons based on element-wise importance scores computed across FFN weight matrices \citep{molchanov2019importance, zhang2022platon, ding2019global}, effectively redistributing model capacity among surviving neurons for smooth compression.

Our contributions in the deployment-constrained, full-cost regime are as follows:

\noindent$\bullet$ We introduce the enlarge-and-prune framework for full-cost evaluation and show that naive enlarge-and-prune pipelines do not consistently outperform training small models from scratch when the enlarged model cost is included.

\noindent$\bullet$ We propose {\our}, an integrated pipeline that unifies enlarged model pretraining, iterative structured pruning, and recovery under a single learning rate schedule. In controlled comparisons on compressing 2.8B models to 1.3B with up to 2T tokens, {\our} consistently outperforms all conventional pipelines overall and improves over the scratch baseline on most reported metrics. These advantages are preserved after downstream supervised fine-tuning on instruction following, math reasoning, and code generation benchmarks.

\noindent$\bullet$ Our analysis reveals that, under a resumed learning rate schedule, intermediate checkpoints---despite weaker absolute performance---provide more favorable starting points for pruning than fully converged checkpoints, and that an enlargement factor of approximately 2.6$\times$ yields the best pruned models in the tested regime.

For rigorous evaluation of the pruning methodology, we isolate our analysis from complementary techniques such as knowledge distillation, though we confirm in separate ablation studies that distillation can be effectively combined with our method.

\section{Background}

\subsection{Iterative Pruning}
Iterative pruning is a parameter reduction technique that alternates between parameter update and removal \citep{zhu2017prune, louizos2017learning}.
During pruning, the method computes importance scores for parameters and gradually removes those less critical.
For a model with parameters $\btheta^{(t)} \in \RR^{N}$, we measure parameter importance by sensitivity \citep{molchanov2019importance}, $\bomega^{(t)} \in \RR^{N}$, which represents the first-order approximation of each parameter's impact on the training loss. See Appendix~\ref{app:sensitivity_score} for the definition.
We then generate a binary mask $\bv^{(t)}$ based on the ranking of $\bomega^{(t)}$, and apply it to obtain the pruned parameters: $\btheta^{(t+1)} = \bv^{(t)} \odot \btheta^{(t)}$.
This progressive approach enables a smooth transition from larger to smaller models.
Note that these works focus on unstructured pruning, while we extend iterative pruning to structured pruning in Section~\ref{method:iterative_pruning}.

\subsection{Structured Pruning of FFN}
\label{sec:structured_pruning}
Transformers \citep{vaswani2017attention}, the mainstream architectures in current generative language models \citep{radford2019language, jiang2023mistral, dubey2024llama, gunter2024apple}, consist of sequential layers containing attention and Feed-Forward Network (FFN) components.
We consider an FFN layer equipped with a gated Sigmoid Linear Unit (SiLU) activation function, denoted as $\sigma(\cdot)$. Given the input $\bx \in \RR^d$, the layer's output $\by \in \RR^d$ is defined as
\begin{align*}
    \by &= \bW_{\rm down}^{\top} (\sigma (\bW_{\rm up} \bx) \odot (\bW_{\rm gate}\bx))   = \sum_{i=1}^{h} \bW_{{\rm down}[i,:]}^{\top} \left(\sigma(\bW_{{\rm up}[i,:]}\bx) \cdot (\bW_{{\rm gate}[i,:]}\bx)\right),
\end{align*}
where $\bW_{\rm up}, \bW_{\rm gate}, \bW_{\rm down} \in \RR^{h \times d}$, $d$ is the input dimension, $h$ is the hidden dimension or the number of neurons, and $\odot$ represents element-wise multiplication.
Structured pruning of FFN width targets the removal of neurons, which corresponds to the rows in $\bW_{\rm up}$, $\bW_{\rm gate}$, and $\bW_{\rm down}$ with the same indices, for example, $\bW_{{\rm up}[i,:]}, \bW_{{\rm gate}[i,:]}, \bW_{{\rm down}[i,:]}$. This is achieved by finding a binary mask $\bbm \in \{0, 1\}^h$ that determines which neurons to retain.
The pruned weight matrices are obtained as:
\begin{equation}
\label{eq:prune_by_mask}
    \begin{split}
    \bW_{\rm up}^{'} = \diag(\bbm) \times \bW_{\rm up}, 
    \bW_{\rm gate}^{'} = \diag(\bbm) \times \bW_{\rm gate}, 
    \bW_{\rm down}^{'} = \diag(\bbm) \times \bW_{\rm down}.
    \end{split}
\end{equation}
After mask application, the zero rows in each matrix are eliminated to obtain compressed weight matrices, resulting in reduced memory consumption and inference latency.

\subsection{Enlarge-and-Prune Framework and Naive Pipeline}
\label{sec:separate_pruning_pipeline}
We introduce the enlarge-and-prune framework to evaluate the end-to-end token efficiency of pruning for producing small models from scratch.
Given an inference budget, practitioners first fix the target small-model architecture for deployment.
The target is then expanded into an enlarged counterpart, for example by increasing the FFN width in each layer, and this enlarged model is pretrained on a corpus.
Next, structured pruning compresses the enlarged model back to the exact target architecture.
A subsequent recovery stage continues pretraining the pruned model to regain performance.
Under the full-cost accounting described in Section~\ref{sec:intro}, this framework enables a controlled comparison between enlarge-and-prune pipelines and direct small-model training.

An immediate implementation under the enlarge-and-prune framework is the naive enlarge-and-prune pipeline, which consists of three distinct stages: enlarged model pretraining, pruning, and pruned model recovery.
From the perspective of learning rate specifically, each of the three stages uses its own independent learning rate schedule \citep{xia2023sheared}.
The learning rate at the $t$-th step through the naive enlarge-and-prune pipeline is
\begin{align*}
  \beta^{(t)} =
  \begin{cases}
    \eta(t; T_l, \eta_1, \eta_2) & \text{if $0 < t \leq T_l$}, \\
    \eta(t - T_l; T_p, \eta_3, \eta_4) & \text{if $ T_l < t \leq T_l + T_p$}, \\
    \eta(t - T_l - T_p; T_r, \eta_5, \eta_6) & \text{if $T_l + T_p < t \leq T$},
  \end{cases}
\end{align*}
where $\eta(\cdot;\cdot)$ is the cosine annealing learning rate schedule \citep{loshchilov2016sgdr} (see Appendix~\ref{app:cosine_lr}), $T$ is the total training steps and $T_l, T_p, T_r$ are the number of enlarged model training steps, pruning steps, pruned model recovery steps, respectively. An example can be found in Figure~\ref{fig:lr_sheared}. In the naive enlarge-and-prune pipeline, we need to tune different sets of peak and end learning rates $\eta_1, \ldots, \eta_6$, which requires considerable effort to find the optimal set.

\subsection{Related Work}

\textbf{Train Large, Then Compress.}
\citet{li2020train} show that pretraining larger models with early stopping and then compressing them is more efficient than training smaller models to convergence.
Our work differs in three respects: we use structured (not unstructured) pruning to preserve inference latency, we prune during pretraining to obtain a general-purpose model rather than during task-specific finetuning, and we target decoder-only architectures where pruning dynamics differ from encoder-only models \citep{liu2019roberta, xia2022structured, xia2023sheared}.

\textbf{Pruning during Pretraining.}
Recent concurrent work also moves pruning into the pretraining phase: \citet{liu2025efficientllm} propose pruning-aware pretraining that extends post-training criteria (LLM-Pruner, SparseGPT) into the pretraining phase for edge LLMs, and \citet{jin2025journey} study sparse pretraining schedules and unify sparse/dense scaling laws, finding that pruning between 25\% and 75\% of training compute is near-optimal for sparse pretraining.
These works focus on unstructured or architecture-search style sparsification, whereas we instantiate structured pruning within a unified cosine annealing schedule that eliminates stage-boundary LR restarts, and we evaluate under full-cost accounting against direct small-model training.

\textbf{Structured Pruning in Post-Training.}
Recent works \citep{ma2023llm, ashkboos2024slicegpt, an2024fluctuation, wei-etal-2024-structured,xu2025lightweight,guo2025slimllm,zhao2025fistapruner} compress pretrained models via structured pruning when checkpoints are off-the-shelf.
Although the goals differ from ours, their pruning criteria can serve as drop-in components within our framework; we demonstrate this with activation-based and learned mask pruning in Section~\ref{sec:extension_pipeline}.

\textbf{Alternative Structured Compression.}
Beyond FFN width reduction, depth pruning \citep{men2024shortgpt, kim2024shortened}, attention head pruning \citep{voita2019analyzing, michel2019sixteen}, and joint pruning of multiple dimensions \citep{krishnakumar2025where} offer complementary strategies.
Our framework is not restricted to any single pruning dimension.
We choose FFN-width as the primary instantiation because FFN layers account for the majority of parameters in decoder-only transformers, making width enlargement the most parameter-efficient strategy \citep{muralidharan2024compact}.
Width pruning also preserves the layer count and attention structure, ensuring the pruned model exactly matches a predetermined deployment architecture.
We provide a matched comparison with attention-head and joint pruning in Section~\ref{sec:prune_attn}.

\textbf{Learning Rate Schedules.}
Recent work shows that the shape of the learning rate schedule, not only its peak value, governs the pretraining loss.
\citet{wen2025understanding} explain Warmup-Stable-Decay (WSD) schedules through a river valley loss landscape, where a high learning rate makes progress along the river and the decay phase descends to the valley floor; \citet{dremov2025training} further analyze the training dynamics of this cooldown stage.
\citet{tissue2024scaling} and \citet{luo2025multipower} predict the loss curve from the entire schedule, attributing the final loss to both the accumulated learning rate and the additional reduction from annealing.
These views are consistent with our resumed-versus-restarted finding (Section~\ref{sec:resume-middle}): restarting the schedule from an annealed checkpoint re-raises the learning rate and discards part of the annealing benefit, whereas resuming a single schedule from an intermediate checkpoint preserves the accumulated progress and keeps a single final decay.

\section{Method}
\label{sec:method}
We now describe IDEA Prune ({\bf I}ntegrate{\bf D} {\bf E}nlarge-{\bf A}nd-{\bf Prune}), which unifies enlarged model pretraining, iterative structured pruning, and recovery into a single training run.

\subsection{Integrated Enlarge-and-Prune Pipeline}
\label{method:integrated_pruning}
We denote the number of enlarged model pretraining, pruning, and pruned model recovery steps as $T_l, T_p, T_r$, respectively. The learning rate for weight matrix update at step $t$ is given by
\begin{align*}
    \alpha^{(t)} = \eta(t; T_l + T_p + T_r, \eta_p, \eta_e),
\end{align*}
where $\eta(\cdot; \cdot)$ is the cosine annealing learning rate schedule defined in \eqref{eq:cosine}.
Unlike the naive enlarge-and-prune pipeline in Section~\ref{sec:separate_pruning_pipeline}, our integrated approach eliminates learning rate warm-up during pruning and recovery stages, thus mitigating potential knowledge loss.

Critically, this learning rate schedule method is not constrained to specific pruning techniques.
We conduct ablation studies to evaluate its effectiveness across diverse approaches, including one-shot random pruning, activation-based pruning, and learned mask pruning (Section~\ref{sec:extension_pipeline}).

Furthermore, the integrated pipeline offers additional flexibility by enabling its application to existing pretrained models with available intermediate checkpoints and historical learning rate schedules, as discussed in Section~\ref{sec:resume-middle}.

\subsection{Iterative Structured Pruning in FFN}
\label{method:iterative_pruning}
To complement the integrated learning rate schedule, we propose a specialized iterative pruning method for FFN width based on the importance score of neurons.
This approach further aligns the pruning and recovery stages by mitigating the suboptimal model initialization inherent in naive enlarge-and-prune pipelines during recovery training.

Concretely, for each element in a weight matrix $\bW^{(t)} \in \RR^{h \times d}$ at the $t$-th step, we denote the importance score as the sensitivity $\bS_{ij}^{(t)}$ defined in \eqref{eq:sensitivity}.
Since the sensitivity is defined on the full dataset, we use the moving average of importance scores on a mini-batch of size $B$ to approximate it as
\begin{align}
\label{eq:moving_avg}
\tilde{\bS}_{ij}^{(t)} = (1 - \lambda)\bT_{ij}^{(t)} + \lambda \tilde{\bS}_{ij}^{(t - 1)},
\end{align}
where
$\bT_{ij}^{(t)} = \Big|\frac{1}{B} \sum_{n=1}^{B} \nabla l_n(\bW_{ij}^{(t)}) \bW_{ij}^{(t)}\Big|,$
$\lambda$ is the moving average coefficient, and $l_n(\cdot)$ is the loss of the $n$-th sample in the mini-batch.
The importance score for the $k$-th neuron is then derived by combining the moving average scores across the three weight matrices $\bW_{{\rm up}}, \bW_{{\rm gate}}, \bW_{{\rm down}}$:
\begin{align}
\label{eq:neuron_impt}
    \bc_k^{(t)} = f_2 \left( f_1(\tilde{\bS}_{{\rm up}[k,:]}^{(t)}), f_1(\tilde{\bS}_{{\rm gate}[k,:]}^{(t)}), f_1(\tilde{\bS}_{{\rm down}[k,:]}^{(t)}) \right),
\end{align}
where $f_1: \RR^{d} \rightarrow \RR$ aggregates element-wise scores into a scalar and $f_2: \RR^3 \rightarrow \RR$ combines across the three matrices. Options of $f_1$ and $f_2$ can be $\max(\cdot), {\rm mean}(\cdot)$.

Finally, we update the pruning mask using a scheduled sparsity by
\begin{align}
\label{eq:mask}
  \bbm_{k}^{(t)} =
  \begin{cases}
    1 & \text{if $\bc_{k}^{(t)}$ is in top $(1 - r(t; T_p))$ fraction of $\bc^{(t)}$}, \\
    0 & \text{otherwise},
  \end{cases}
\end{align}
where $r(t; T_p) \in [0, R]$ is the sparsity (fraction of neurons removed) at step $t$, defined as a cubically increasing function in Appendix~\ref{appendix:cubic_sparsity_schedule} that reaches the target sparsity $R$. The full {\our} algorithm is summarized in Algorithm~\ref{alg:integrated_pipeline} (Appendix~\ref{appendix:algorithm}).

\section{Experiments}
\label{sec:pretraining-pruning}
We investigate the effectiveness of {\our} compared to the naive pipeline with baseline pruning methods, and study the token efficiency of the enlarge-and-prune pipeline compared to training small models from scratch.
We further fine-tune all pretrained models with an identical recipe to examine whether these advantages are preserved in downstream tasks.

\noindent {\bf Experimental Setup.}
We choose a 1.3B target model and a 2.8B enlarged model differing only in FFN hidden dimension (Appendix~\ref{appendix:model_arch}).
All models are trained on DCLM \citep{li2024datacomp} (4T tokens) using AdamW \citep{loshchilov2017decoupled} with sequence length 4096 and batch size 1024 on Google Cloud v5p TPUs using AXLearn \citep{axlearn2024}.
We compare against (i) training 1.3B from scratch with 1T and 2T tokens, and (ii) naive enlarge-and-prune pipelines using OSRP, Minitron \citep{muralidharan2024compact}, and Sheared LLaMA \citep{xia2023sheared} (Appendix~\ref{appendix:baselines}).
For {\our}, we allocate 1T, 0.5T, and 0.5T tokens to enlarged pretraining, pruning, and recovery under a single cosine annealing schedule.
We report perplexity on OpenWebText \citep{Gokaslan2019OpenWeb}, 0-shot accuracy on Arc-C \citep{allenai2018arc}, SciQA \citep{welbl2017crowdsourcing}, and HellaSwag \citep{zellers2019hellaswag}, 1-shot on TriviaQA \citep{joshi2017triviaqa}, and 5-shot on MMLU \citep{hendrycks2020measuring}.
Since the pruned model has the same architecture as the scratch baseline, inference latency and memory are identical; see Appendix~\ref{appendix:hyperparams} and \ref{appendix:compute} for full details.

\noindent {\bf Main Results.}
As shown in Table~\ref{tab:pretraining-pruning}, {\our} improves over all naive enlarge-and-prune pipelines.
The improvement on OpenWebText and MMLU is the most significant: our method attains an OpenWebText perplexity of 8.88, outperforming the previous best of 8.96 achieved by Sheared LLaMA, and advances MMLU accuracy to 46.4\%, compared to 31.4--33.4\% for naive methods.
However, different pipelines show minimal difference on reading comprehension and commonsense tasks: Arc-C, SciQA, HellaSwag, and TriviaQA.
This is likely because performance saturates on these tasks at this model scale.
As we increase the training tokens from 1T to 2T for the 1.3B model, these tasks are not significantly improved, but MMLU continues to improve.
Results on Arc-Easy, WinoGrande, and CommonsenseQA show the same pattern (Appendix~\ref{appendix:more_benchmarks}).

\begin{table}[htb!]
\vspace{-0.175in}
\begin{center}
\begin{small}
\caption{\label{tab:pretraining-pruning} Evaluation of models trained from scratch or through enlarge-and-prune pipelines. We train a 1.3B model for 2T tokens as the baseline. We apply naive pipelines with different pruning methods to prune a 2.8B model to 1.3B with 2T tokens. We report the perplexity($\downarrow$) of OpenWebText and accuracy($\uparrow$) of other benchmarks. The best results are in {\bf bold}.}
\resizebox{\linewidth}{!}{%
\begin{tabular}{c|cccccc}
\toprule
{\bf Method} & {\bf OWT} $\downarrow$ & {\bf Arc-C} $\uparrow$ & {\bf SciQA} $\uparrow$ & {\bf HellaSwag} $\uparrow$ & {\bf TriviaQA} $\uparrow$ & {\bf MMLU} $\uparrow$ \\
\midrule
{1.3B-1T from scratch} & {9.10} & {39.3} & {93.6} & {52.8}  & {30.3} & {28.9}\\
{2.8B-1T from scratch} & {8.12} & {44.5} & {94.2} & {56.9} & {37.3} & {50.8}\\
\midrule
{1.3B-2T from scratch} & {8.95} & {\bf 39.4} & {93.4} & {53.6} & {30.5} & {45.7}\\
\midrule
{OSRP-2T (1.3B)} & {8.98} & {38.9} & {92.9} & {53.4} & {29.6} & {32.5} \\
{Minitron-2T (1.3B)}  & {8.97} & {38.7} & {\bf 94.6} & {53.6} & {30.7} & {31.4} \\
{Sheared LLaMA-2T (1.3B)} & {8.96} & {\bf 39.4} & {94.0} & {53.6} & {30.1} & {33.4} \\
{\bf {\our}-2T (1.3B)}  & {\bf 8.88} & {39.0} & {\bf 94.6} & {\bf 54.0} & {\bf 31.1} & {\bf 46.4} \\
\bottomrule
\end{tabular}%
}
\end{small}
\end{center}
\vspace{-0.175in}
\end{table}

Compared to the 1.3B-2T scratch baseline, {\our} is slightly better or comparable on most metrics (Table~\ref{tab:pretraining-pruning}), confirming that carefully designed enlarge-and-prune pipelines can match or exceed direct training under the same token budget.

For controlled comparison against naive pipelines, the main 2.8B$\to$1.3B experiment uses a matched 50/50 allocation (1T for enlarged model training, 1T for pruning and recovery).
Proxy-scale ablations in Section~\ref{sec:ablate_pruning_schedule} suggest that more recovery-heavy allocations (e.g., 30/70) can further improve {\our}, but we do not claim the main-scale result is fully tuned.
Even under this matched configuration, {\our} achieves improvements over both training small models from scratch and naive pipelines.

\noindent {\bf Downstream Fine-Tuning.}
\label{sec:sft}
To examine whether the pretraining-stage gains persist after post-training, we fine-tune the pretrained checkpoints of Table~\ref{tab:pretraining-pruning} with the same supervised fine-tuning (SFT) recipe.
Every checkpoint is fine-tuned on Dolci-Instruct-SFT \citep{olmo2025olmo3} (approximately 2M examples) for 2 epochs, with a cosine learning rate schedule from $2 \times 10^{-5}$ to $1 \times 10^{-8}$ and a batch size of 128.
The data, recipe, and compute are identical for every row; only the starting checkpoint differs.
We evaluate on MMLU \citep{hendrycks2020measuring}, GSM8K \citep{cobbe2021training}, IFEval \citep{zhou2023instruction}, and HumanEval \citep{chen2021evaluating}.

Table~\ref{tab:sft} shows that {\our} is the best 1.3B model on every benchmark under the same total pretraining budget: it matches or exceeds the scratch baseline and outperforms every naive pipeline, with the largest margin on HumanEval (+3.1 over the scratch baseline and +4.9 to +7.3 over naive pipelines).
The ordering of methods before SFT ({\our} $\geq$ scratch $>$ naive pipelines) is preserved after SFT, indicating that the pretraining-stage gains do not wash out under fine-tuning.

\begin{table}[htb!]
\centering
\caption{\label{tab:sft} Results after SFT. Each row is fine-tuned from the corresponding pretrained checkpoint in Table~\ref{tab:pretraining-pruning} with an identical recipe. IFEval reports prompt-level (P) and instruction-level (I) accuracy under loose and strict criteria. The best 1.3B results are in {\bf bold}.}
\begin{small}
\resizebox{\linewidth}{!}{%
\begin{tabular}{c|ccccccc}
\toprule
\multirow{2}{*}{\bf Initial Checkpoint} & \multirow{2}{*}{\bf MMLU} & \multirow{2}{*}{\bf GSM8K} & \multicolumn{4}{c}{\bf IFEval} & \multirow{2}{*}{\bf HumanEval} \\
~ & ~ & ~ & {P-loose} & {P-strict} & {I-loose} & {I-strict} & ~ \\
\midrule
{2.8B-1T from scratch} & {51.4} & {41.2} & {71.2} & {67.7} & {78.9} & {76.1} & {31.4} \\
\midrule
{1.3B-2T from scratch} & {45.1} & {32.1} & {\bf 66.5} & {\bf 62.5} & {74.7} & {71.6} & {24.3} \\
\midrule
{OSRP-2T (1.3B)} & {42.2} & {28.4} & {64.1} & {60.3} & {73.4} & {70.1} & {20.1} \\
{Minitron-2T (1.3B)} & {43.8} & {32.0} & {63.2} & {59.7} & {73.3} & {70.4} & {20.1} \\
{Sheared LLaMA-2T (1.3B)} & {44.2} & {30.3} & {63.8} & {61.5} & {75.0} & {71.7} & {22.5} \\
{\our}-2T (1.3B) & {\bf 46.0} & {\bf 33.0} & {\bf 66.5} & {\bf 62.5} & {\bf 76.0} & {\bf 72.8} & {\bf 27.4} \\
\bottomrule
\end{tabular}%
}
\end{small}
\end{table}

\section{Analysis and Extensions}

\subsection{Analysis of Initialization and Learning Rate}
\label{sec:resume-middle}
If we only consider the pruning and recovery stages, the major differences of the integrated and naive enlarge-and-prune pipelines are twofold: the {\it initial weights} of the model before pruning and the {\it learning rate schedule}.

{\bf Initialization.} At the beginning of the pruning stage, naive enlarge-and-prune pipelines initialize the model by the last checkpoint of an enlarged model trained with, e.g., 1T tokens.
In contrast, {\our} loads the model weights that are equivalent to the intermediate checkpoint of an enlarged model with a larger token budget, e.g., 2T tokens if we did not prune the enlarged model and continued to fully train the enlarged model.
To ablate the impact of the initialization, we initialize the model by two checkpoints: the checkpoint of the 2.8B-2T model at the 1T-th step and the checkpoint of the 2.8B-1T at its last step.

{\bf Learning rate schedule.} In {\our}, we resume the learning rate schedule of the 2.8B-2T model at the 1T point, which provides a relatively small learning rate continuation, while naive enlarge-and-prune pipelines restart the cosine learning rate schedule with a linear warmup.
We formalize these approaches into two learning rate schedule types: first, the {\it resumed} learning rate schedule, corresponding to the integrated enlarge-and-prune pipeline, mathematically represented by $\eta(t+T_l; T_l + T_p + T_r, \eta_p, \eta_e)$, where the learning rate continues from a previous training stage; and second, the {\it restarted} learning rate schedule, corresponding to the naive enlarge-and-prune pipeline, represented by $\eta(t; T_p + T_r, \eta_p, \eta_e)$, which initiates a new learning rate schedule from the beginning. We fix $\eta_p, \eta_e$ in each setting.

{\bf Discussion 1.} We present the results in Table~\ref{tab:ablation_full}.
First, both the resumed learning rate schedule and the use of intermediate checkpoints are crucial components; omitting either leads to substantial degradation in MMLU.
Intriguingly, we find that the learning rate schedule has a greater impact than initialization, with the resumed schedule consistently outperforming the restarted approach across different initialization methods.

{\bf Discussion 2.} Contrary to conventional wisdom, our results challenge the presumption that better initialization directly translates to better pruned model performance.
Most notably in the resumed learning schedule, we discover that an intermediate checkpoint, despite showing lower performance than the final checkpoint, paradoxically provides a more favorable starting point for pruning, resulting in higher pruned model performance.
This counterintuitive result stems from the learning dynamics: the final checkpoint's convergence at a low learning rate creates a mismatch with the resumed schedule's relatively higher learning rate, effectively mimicking a restarted schedule but with smaller peak learning rates.
These findings emphasize the critical importance of aligning initialization with learning rate schedules in enlarge-and-prune pipelines.

\begin{table}[htb!]
\begin{center}
\begin{small}
\caption{\label{tab:ablation_full}
Ablation of the initialization and the learning rate schedule in enlarge-and-prune pipelines. 2.8B-2T@1T indicates the checkpoint at the 1T-token step of a 2.8B training with a 2T total token budget.
Pruned model size: 1.3B.
}
\resizebox{\linewidth}{!}{%
\begin{tabular}{cc|cccccc}
\toprule
{\bf Model} & {\bf LR Schedule} & {\bf OWT}$\downarrow$ & {\bf Arc-C}$\uparrow$ & {\bf SciQA}$\uparrow$ & {\bf HellaSwag}$\uparrow$ & {\bf TriviaQA}$\uparrow$ & {\bf MMLU}$\uparrow$ \\
\midrule
{2.8B-2T@1T} & {-} & {9.99} & {38.7} & {93.5} & {51.6} & {24.8} & {26.0} \\
\cmidrule(lr){1-8}
\multirow{2}{*}{{\parbox{2.5cm}{\centering 1.3B pruned from 2.8B-2T@1T}}} & {Resumed} & {\bf 8.88} & {39.0} & {94.6} & {\bf 54.0} & {\bf 31.1} & {\bf 46.4} \\
~ & {Restarted} & {8.94} & {39.7} & {93.7} & {54.5} & {30.1} & {37.0} \\
\midrule
{2.8B-1T@1T} & {-} & {8.12} & {44.5} & {94.2} & {56.9} & {37.3} & {50.8} \\
\cmidrule(lr){1-8}
\multirow{2}{*}{{\parbox{2.5cm}{\centering 1.3B pruned from 2.8B-1T@1T}}} & {Resumed} & {8.89} & {39.3} & {93.5} & {53.6} & {31.8} & {38.2} \\
~ & {Restarted} & {8.95} & {39.9} & {92.9} & {53.3} & {31.0} & {36.8} \\
\bottomrule
\end{tabular}%
}
\end{small}
\end{center}
\end{table}

\subsection{Analysis of Enlarged Model Size}
\label{sec:model_size}

Unlike pruning existing enlarged models, enlarge-and-prune pipelines have more freedom on the choice of the enlarged model size.
Therefore, we investigate the impact of enlarged model size on performance through the enlarge-and-prune pipelines.
Starting with a 300M parameter baseline model featuring an FFN layer with $3 \times 1024$ hidden dimensions (detailed architecture in Appendix~\ref{appendix:model_arch}), we explore enlarged models by incrementally increasing FFN width to $4 \times$, $6 \times$, $8 \times$, $12 \times$, $16 \times$ of 1024, while keeping other model parameters constant.
We conduct {\our} using 500B tokens from DCLM, allocating 250B tokens to the enlarged model and 250B tokens to pruning and recovery.

\begin{wrapfigure}{r}{0.35\linewidth}
\vspace{-12pt}
\centering
\includegraphics[width=\linewidth]{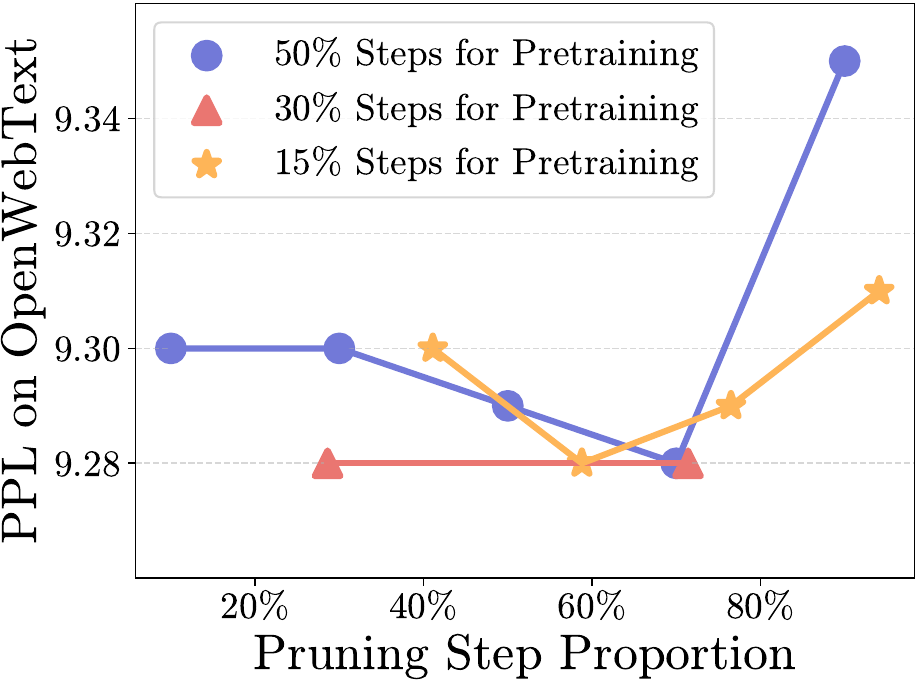}
\vspace{-12pt}
\caption{\label{fig:pruning_schedule}Pruning schedule.}
\vspace{-10pt}
\end{wrapfigure}

Figure~\ref{fig:ppl_size} illustrates our experiment results. Notably, integrated pipelines with all enlarged model sizes significantly outperform the training from scratch baseline.
This robust performance across a wide range of model sizes demonstrates the flexibility of our integrated enlarge-and-prune pipeline and minimal need for model size tuning.
Additionally, the results reveal an important trade-off between model capacity and pruning efficiency.
Models with smaller FFN width enlargement factors (close to 1.3$\times$) suffer from insufficient capacity to learn rich representations, while extremely large models (approaching 5.3$\times$) face challenges of pruning degradation.
Based on these observations, we identify 2.6$\times$ as the optimal FFN width enlargement factor in the tested regime.

\subsection{Robustness of Hyperparameters}
\label{sec:hyper_1}

We study how the pruning schedule, importance score combination functions, and the moving average coefficient affect {\our}; all are robust and require little tuning.

{\bf Pruning Schedule.}
\label{sec:ablate_pruning_schedule}
Using a 400B token budget, we vary pretraining step proportions at 15\%, 30\%, and 50\%, and examine different pruning steps $T_p$ used in $r(t;T_p)$.
Figure~\ref{fig:pruning_schedule} reveals robust performance across a wide range of steps.
For instance, with 50\% steps in pretraining, valid pruning steps extend from 10\% to 70\%.
The analysis suggests an optimal pruning proportion around 70\%, indicating potential for further optimization beyond the 50/50 split used in our main experiments.
Other hyperparameters---importance score combination ($f_1, f_2$) and moving average coefficient ($\lambda$)---are also robust; see Appendix~\ref{appendix:hyperparam_ablation} for details.

\subsection{Scaling Property}
\label{sec:scaling}
\begin{figure}[htb!]
\vspace{-0.1in}
\centering
\begin{subfigure}{0.45\linewidth}
    \centering
    \includegraphics[width=0.9\linewidth]{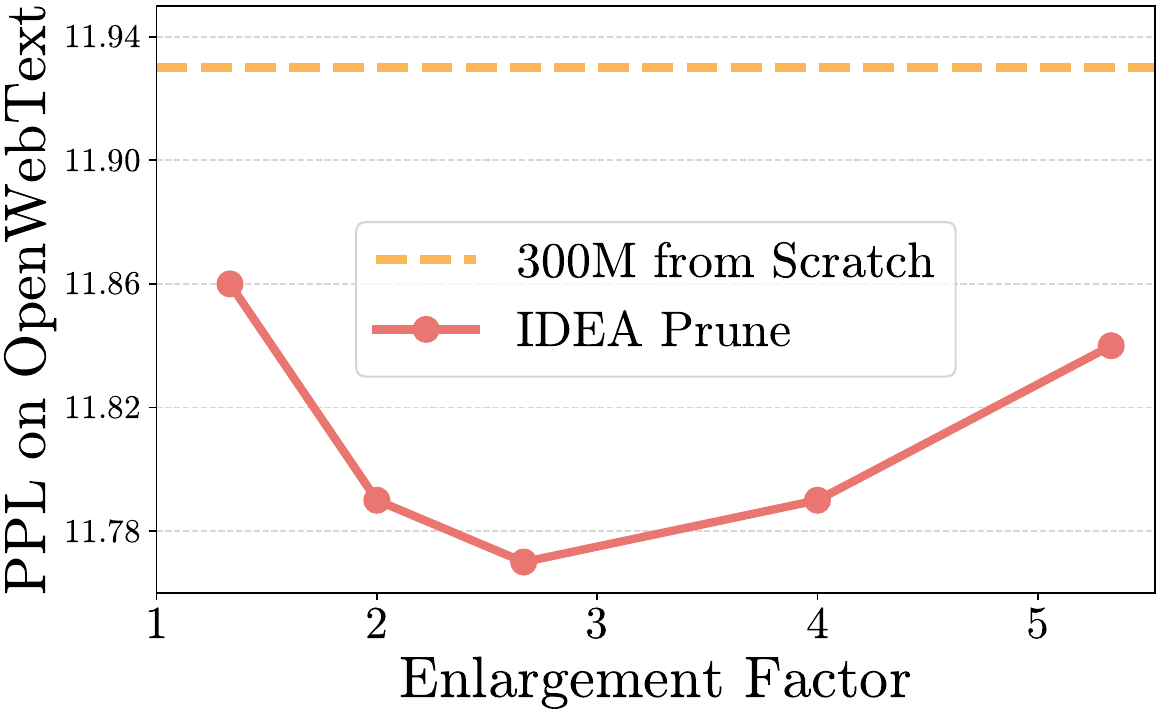}
    \caption{\label{fig:ppl_size} Enlargement factor}
\end{subfigure}
\hspace{0.3in}
\begin{subfigure}{0.45\linewidth}
    \centering
    \includegraphics[width=0.9\linewidth]{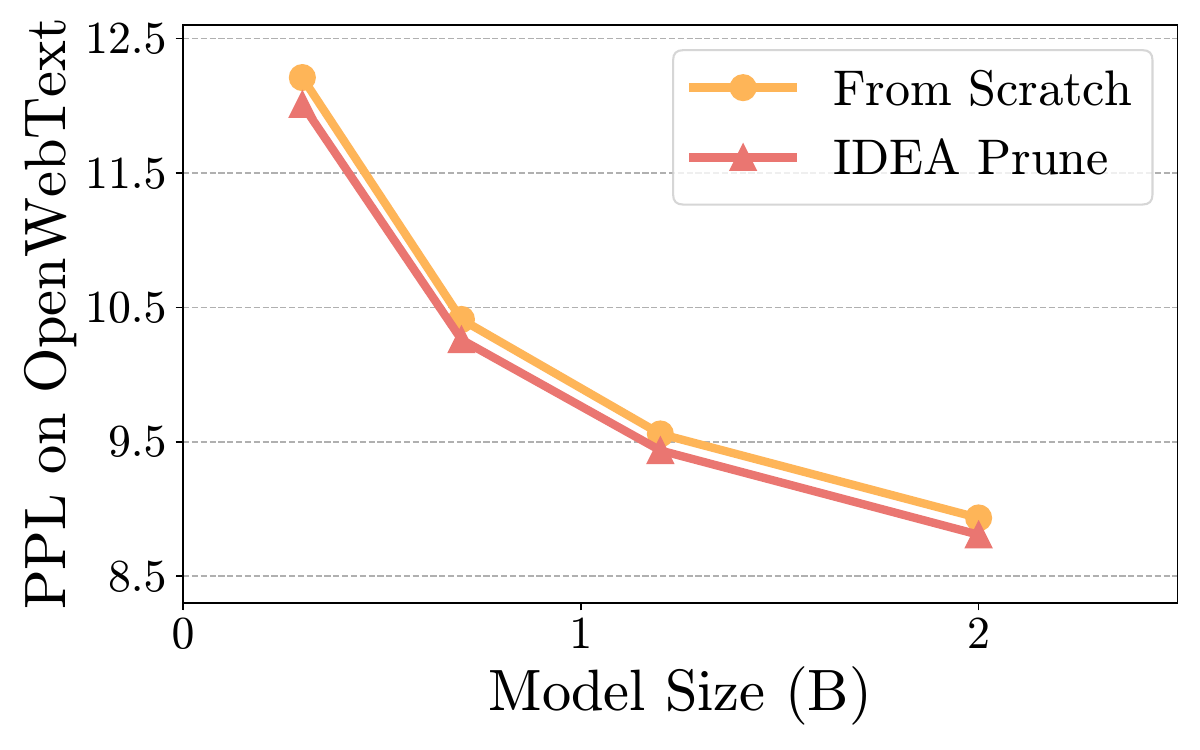}
    \caption{\label{fig:ppl_scaling} Target model size}
\end{subfigure}
\vspace{-0.05in}
\caption{{\bf Left}: Ablation of the enlarged model size. {\bf Right}: Scaling property of {\our}.}
\vspace{-0.1in}
\end{figure}

We study whether the advantage of {\our} over training from scratch holds across different target model sizes.
We choose target sizes of 300M, 700M, 1.2B, and 2B with corresponding enlarged models of 680M, 1.5B, 2.8B, and 4.2B.
Each model is trained with 300B tokens from DCLM.
As shown in Figure~\ref{fig:ppl_scaling}, {\our} outperforms the scratch baseline in perplexity across all target sizes within the on-device model range.

\subsection{Extension to Attention-Head and Joint Pruning}
\label{sec:prune_attn}
{\our} is not restricted to FFN-width pruning.
To test this, we apply {\our} to attention-head pruning and to joint FFN-and-attention pruning at the 300M target scale, with a 300B-token budget for all models.
For attention-only enlargement, we double the number of attention heads and KV heads while keeping the per-head dimension fixed, yielding enlarged models of 380M and 450M parameters.
For joint enlargement, we scale both attention and FFN dimensions, yielding a 750M enlarged model (see Appendix~\ref{appendix:model_arch} for configurations).
All models are pruned back to the 300M target architecture.

\begin{table}[htb!]
\centering
\caption{\label{tab:prune-attn} Extension of {\our} to attention-head and joint pruning at the 300M target scale. All models are trained with a 300B-token budget. ``$-$'' denotes the small model from scratch baseline.}
\begin{small}
\begin{tabular}{c|ccccc}
\toprule
{Enlarged Dims} & {--} & {Attn (2$\times$)} & {Attn (3$\times$)} & {FFN} & {Attn\&FFN} \\
{Enlarged Size} & {300M} & {380M} & {450M} & {680M} & {750M} \\
\midrule
{OWT Perplexity} & {12.21} & {12.18} & {12.20} & {11.99} & {11.98} \\
\bottomrule
\end{tabular}
\end{small}
\end{table}

As shown in Table~\ref{tab:prune-attn}, attention-only pruning provides a small improvement over training from scratch (12.18 vs.\ 12.21), while FFN-only pruning yields a substantially larger gain.
Joint FFN-and-attention pruning (11.98) performs comparably to FFN-only, suggesting that the marginal benefit of attention pruning is small once FFN pruning is applied.
These results support our choice of FFN-width as the primary pruning target in the tested 300M setting.

\subsection{Combination with Knowledge Distillation}
\label{appendix:kd}

\begin{wraptable}{r}{0.45\linewidth}
\vspace{-24pt}
\centering
\caption{\label{tab:distill} KD combined with {\our}. Teacher: 2.8B-2T.}
\vspace{-6pt}
\begin{small}
\begin{tabular}{c|c|cc}
\toprule
\multirow{2}{*}{} & {Prune} & {1.3B-1T} & {Prune} \\
{} & {w/o KD} & {w/ KD} & {w/ KD} \\
\midrule
{OWT} & 8.88 & 8.95 & 8.86 \\
{Arc-C} & 39.0 & 39.6 & 39.8 \\
{SciQA} & 94.6 & 93.9 & 93.8 \\
{HSwag} & 54.0 & 52.2 & 52.7 \\
{TQA} & 31.1 & 31.7 & 32.0 \\
{MMLU} & 46.4 & 44.5 & 46.6 \\
\bottomrule
\end{tabular}
\end{small}
\vspace{-10pt}
\end{wraptable}

Knowledge distillation (KD) \citep{kim2016sequence, hinton2015distilling} is complementary to pruning and is often combined with it in practice \citep{team2024gemini, liang2023homodistil,gunter2024apple, li2023losparse}.
Under our full-cost framework, we study whether KD can further improve the integrated enlarge-and-prune pipeline when the teacher model cost is included in the total token budget.
We train a 2.8B-2T model as the teacher and establish a KD baseline by training a 1.3B model from scratch over 1T tokens with KD.
In our pipeline, we prune and train the 2.8B-2T@1T intermediate checkpoint using a resumed cosine learning rate schedule for 1T tokens, with the 2.8B-2T model as the KD teacher.
As shown in Table~\ref{tab:distill}, pruning with distillation outperforms the pruning baseline across most benchmarks and surpasses or is on par with the KD baseline on all benchmarks, particularly on OpenWebText and MMLU.

\subsection{Extension to Other Pruning Methods}
\label{sec:extension_pipeline}
To show the integrated pipeline is not tied to iterative pruning, we apply the integrated learning rate schedule to activation-based pruning (one-shot pruning of the 2.8B-2T@1T checkpoint using activation magnitudes; Appendix~\ref{appendix:minitron_importance}) and learned mask pruning (mask learning from the same checkpoint, with a resumed LR schedule for weights and a fully decaying schedule for mask parameters). Recovery uses the resumed schedule of Section~\ref{sec:resume-middle}.

Table~\ref{tab:int_sep_full} shows the integrated pipeline consistently beats its naive counterpart across all pruning methods, with the largest gain on MMLU (9--13 points). This confirms that the LR restart in naive pipelines causes substantial knowledge loss regardless of criterion, and that any criterion can benefit from replacing the naive LR restarts with our unified schedule.

\begin{table}[htb!]
\vspace{-0.16in}
\begin{center}
\begin{small}
\caption{\label{tab:int_sep_full}
Extension of the integrated pipeline to activation-based pruning and learned mask pruning, alongside OSRP. We prune a 2.8B model to 1.3B with 2T tokens. The integrated pipeline consistently outperforms the naive counterpart for every pruning criterion.
}
\begin{tabular}{cc|cccccc}
\toprule
\multicolumn{2}{c|}{\bf Method} & {\bf OWT}$\downarrow$ & {\bf Arc-C}$\uparrow$ & {\bf SciQA}$\uparrow$ & {\bf HellaSwag}$\uparrow$ & {\bf TriviaQA}$\uparrow$ & {\bf MMLU}$\uparrow$ \\
\midrule
\multirow{2}*{OSRP} & {Naive} & {8.980} & {38.9} & {92.9} & {53.4} & {29.6} & {32.5} \\
~ & {Integrated} & {8.926} & {41.1} & {93.4} & {53.8} & {30.1} & {42.9} \\
\midrule
\multirow{2}*{Activation-based} & {Naive} & {8.973} & {38.7} & {94.6} & {53.6} & {30.7} & {31.4} \\
~ & {Integrated} & {8.916} & {39.8} & {94.0} & {53.6} & {30.8} & {43.9} \\
\midrule
\multirow{2}*{Learned mask} & {Naive} & {8.960} & {39.4} & {94.0} & {53.6} & {30.1} & {33.4} \\
{~} & {Integrated} & {8.892} & {43.2} & {93.9} & {54.0} & {30.5} & {42.4}\\
\bottomrule
\end{tabular}
\end{small}
\end{center}
\vspace{-0.16in}
\end{table}

\section{Discussion and Limitations}
\label{sec:limitations}

\begin{figure}[htb!]
\centering
\begin{subfigure}{0.42\linewidth}
    \centering
    \includegraphics[width=1.0\linewidth]{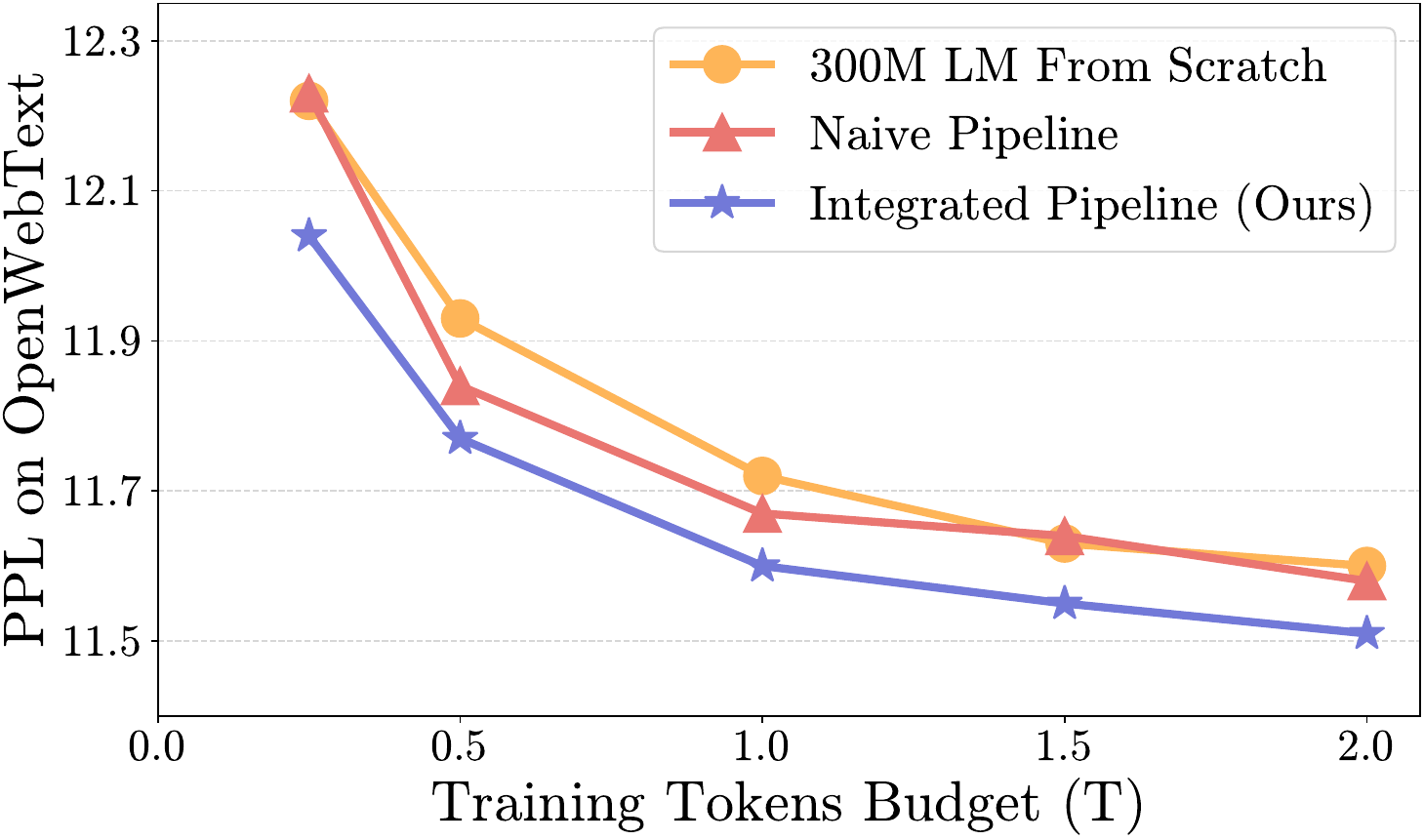}
    \caption{\label{fig:ppl_token} Token Budget}
\end{subfigure}
\begin{subfigure}{0.42\linewidth}
    \centering
    \includegraphics[width=1.0\linewidth]{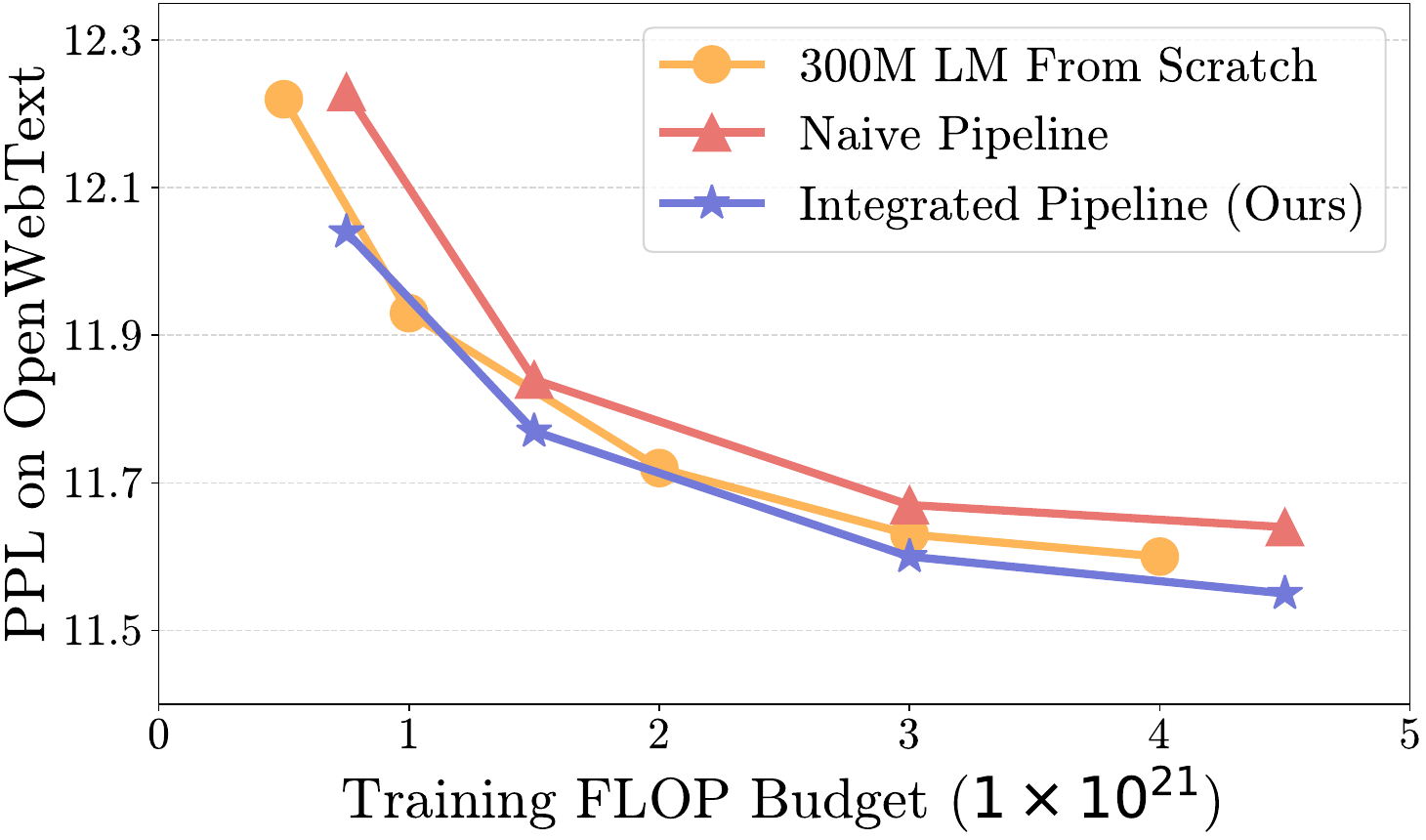}
    \caption{\label{fig:ppl_flops} FLOPs Budget}
\end{subfigure}
\caption{\label{fig:ppl_token_vs_flops}
{\bf Left}: Perplexity on OpenWebText of 300M model trained with multiple training token budgets from 0.25T to 2T.
{\bf Right}: FLOPs of each point from the left figure.}
\end{figure}

{\bf Tokens or FLOPs?}
Our main comparisons fix the training token budget, since tokens are the practical bottleneck in our regime.
Pruning pipelines intrinsically consume more FLOPs than training the target-sized small model given the same training tokens.
In wall-clock time, however, the step time of the 1.3B, 2.8B, and pruning runs is nearly identical in our setup (Appendix~\ref{appendix:compute}).
As shown in Figure~\ref{fig:ppl_token_vs_flops}, even under matched FLOPs, {\our} still outperforms naive pipelines at all budgets and surpasses the scratch baseline at higher budgets, which is practical when practitioners have sufficient computational resources.

{\bf Significance and Reliability.}
The gains of {\our} over training from scratch are modest in absolute terms but meaningful in a regime where small-model performance saturates quickly.
To verify that these gains are not due to noise: (1) final-stage checkpoints of the 1.3B scratch model show minimal variance (Table~\ref{tab:last_3_ckpt_results}), used here as a proxy for convergence stability in the 2.8B$\to$1.3B regime, consistent with prior findings on convergence stability \citep{van2025polypythias}; and (2) proxy-scale multi-seed runs (300M target, 3 seeds, 300B tokens) yield perplexities of 12.01/11.99/11.99 for {\our} and 12.22/12.21/12.23 for the scratch baseline, confirming negligible variance.

{\bf Limitations inherent to large-scale pretraining.}
Pretraining experiments at the scales studied in this paper are computationally expensive, which inevitably constrains the scope of our empirical investigation:
(1) the matched comparison between FFN-width and attention-head pruning is conducted at the 300M proxy scale (Section~\ref{sec:prune_attn}) but not at the main 2.8B$\to$1.3B scale;
(2) scaling validation covers four target sizes up to 2B (Section~\ref{sec:scaling}), but not beyond;
(3) the main experiments are single-run, though proxy multi-seed experiments and convergence analysis (Table~\ref{tab:last_3_ckpt_results}) suggest negligible variance;
(4) larger source models are beyond our resources, as a single 2.8B$\to$1.3B run already costs approximately $3.8 \times 10^{22}$ FLOPs (Appendix~\ref{appendix:compute}).

\bibliographystyle{abbrvnat}
\bibliography{example_paper}

\appendix

\section{Model Optimization}
\subsection{AdamW Optimizer}
\label{app:adam}
We use the AdamW optimizer \citep{loshchilov2017decoupled} with weight decay.
Given a weight matrix $\bW^{(t-1)}$ at the $t$-th step, the weight update is
\begin{equation}
\label{eq:adam}
    \bW^{(t)} = (1 - \gamma^{(t)} \lambda_{\text{wd}}) \bW^{(t-1)} - \gamma^{(t)} \tilde{\bG}^{(t)},
\end{equation}
where $\gamma^{(t)}$ is the learning rate, $\lambda_{\text{wd}}$ is the decoupled weight decay coefficient, and $\tilde{\bG}^{(t)}$ is calculated as
\begin{equation*}
    \tilde{\bG}^{(t)} = \frac{\tilde{\bM}^{(t)}}{\sqrt{\tilde{\bV}^{(t)}} + \epsilon},
\end{equation*}
where $\epsilon$ is a hyperparameter, and $\tilde{\bM}^{(t)}, \tilde{\bV}^{(t)}$ are determined by $\bG^{(t-1)}$, the gradient of $\bW^{(t-1)}$, and the optimization states $\bM^{(t-1)}, \bV^{(t-1)}$. Specifically, we update $\bM^{(t-1)}, \bV^{(t-1)}$ by
\begin{align*}
    \bM^{(t)} &= \beta_1 \bM^{(t-1)} + (1 - \beta_1) \bG^{(t-1)}, \\
    \bV^{(t)} &= \beta_2 \bV^{(t-1)} + (1 - \beta_2) \bG^{(t-1)} \odot \bG^{(t-1)},
\end{align*}
where $\beta_1$ and $\beta_2$ are hyperparameters.
Then, $\tilde{\bM}^{(t)}$ and $\tilde{\bV}^{(t)}$ are calculated as
\begin{align*}
    \tilde{\bM}^{(t)} &= \bM^{(t)} / (1 - \beta_1^t), \\
    \tilde{\bV}^{(t)} &= \bV^{(t)} / (1 - \beta_2^t).
\end{align*}

\subsection{Cosine Annealing Learning Rate Schedule}
\label{app:cosine_lr}
The learning rate $\gamma^{(t)}$ in \eqref{eq:adam} is scheduled by a cosine annealing with linear warm-up \citep{loshchilov2016sgdr}. Given a total of $T$ training steps, the schedule is
\begin{align}
  \eta(t; T, \eta_p, \eta_e) =
  \begin{cases}
    \frac{t}{T_0} \eta_p & \text{if $0 < t \leq T_0$}, \\
    c_1 \cos\left(\frac{t - T_0}{T - T_0} \pi\right) + c_2 & \text{if $T_0 < t \leq T$} ,
  \end{cases}
\label{eq:cosine}
\end{align}
where $c_1=\frac{1}{2}(\eta_p - \eta_e), c_2=\frac{1}{2} (\eta_p + \eta_e)$, $T_0$ represents the number of warmup steps, and $\eta_p$ and $\eta_e$ are the peak and end learning rates, respectively.

\section{IDEA Prune Algorithm}
\label{appendix:algorithm}

\begin{algorithm}
    \caption{{\our}\label{alg:integrated_pipeline}}
    \begin{algorithmic}[1]
        \INPUT{Sparsity schedule $r(t; T_p)$, learning rate schedule $\eta(t; T, \eta_p, \eta_e)$, pruning interval $K$, initial FFN width $h$.}
        \STATE{Randomly initialize model weights $\bW^{(0)}$.}
        \STATE{Zero initialize approximate importance scores $\tilde{\bS}^{(0)}$; set living neurons $\cA \leftarrow \{1, \ldots, h\}$.}
        \FOR {t = $1$ to $T$}
            \STATE{Compute gradients $\nabla l_i(\bW^{(t-1)})$ on a mini batch.}
            \STATE{Compute learning rate $\alpha^{(t)} \leftarrow \eta(t; T, \eta_p, \eta_e)$.}
            \STATE{Update weights $\bW^{(t)}$ following \eqref{eq:adam} with $\gamma^{(t)} = \alpha^{(t)}$.}
            \STATE{Update importance scores $\tilde{\bS}^{(t)}$ following \eqref{eq:moving_avg}.}
            \IF{$t \bmod K = 0$ and $h - |\cA| < \lfloor r(t; T_p)\, h \rfloor$}
                \STATE{Compute neuron importance $\bc_k^{(t)}$ for $k \in \cA$ following \eqref{eq:neuron_impt}.}
                \STATE{$k^* \leftarrow \argmin_{k \in \cA} \bc_k^{(t)}$; $\cA \leftarrow \cA \setminus \{k^*\}$.}
                \STATE{Permanently remove neuron $k^*$: set $\bbm_{k^*} = 0$ in \eqref{eq:prune_by_mask} and drop the $k^*$-th rows of $\bW_{\rm up}, \bW_{\rm gate}, \bW_{\rm down}$.}
            \ENDIF
        \ENDFOR
        \OUTPUT{$\bW^{(T)}$}
    \end{algorithmic}
\end{algorithm}

Pruning and recovery are two phases of the same training loop and differ only in whether neuron removal (lines 8--12) fires.
During enlarged pretraining ($t \leq T_l$), $r(t; T_p) = 0$, so no neuron is removed.
During pruning, one neuron is removed every $K$ steps, so that the number of removed neurons tracks the cubic sparsity schedule $r(t; T_p)$ (Appendix~\ref{appendix:cubic_sparsity_schedule}); the discrete one-neuron-every-$K$-steps cadence realizes this continuous schedule given that the neuron count is an integer.
During recovery, the target width has been reached, so no neuron is removed.
The weight update runs identically in all three phases, and the optimizer state and the single cosine learning rate schedule carry through without any restart.
Removal is permanent rather than a soft mask: a removed neuron no longer exists in the compacted dense weight matrices, so it cannot regain a score or re-enter later.
In the main experiment, enlarged pretraining, pruning, and recovery span 1T, 0.5T, and 0.5T tokens, respectively.

\section{Iterative Pruning}
\subsection{Sensitivity Score}
\label{app:sensitivity_score}
For a model with parameters $\btheta^{(t)} \in \RR^{N}$, we measure parameter importance by sensitivity \citep{molchanov2019importance}:
\begin{align}
\label{eq:sensitivity}
\bomega^{(t)} = \left| \nabla \cL(\btheta^{(t)}) \odot \btheta^{(t)} \right|,
\end{align}
where $\cL(\cdot)$ is the loss function.
This score represents the first-order approximation of each parameter's impact on the loss: $\bomega^{(t)} \approx \left| \cL(\btheta^{(t)}) - \cL(\btheta_{-k}^{(t)}) \right|$, where $\btheta_{-k}^{(t)} = \left(\theta_1, \ldots, \theta_{k-1}, 0, \theta_{k+1}, \ldots, \theta_N \right)$.

\subsection{Cubic Sparsity Schedule in Iterative Pruning}
\label{appendix:cubic_sparsity_schedule}
The sparsity $r(t; T_p)$ (fraction of neurons removed) at the $t$-th step increases cubically from 0 to the target sparsity $R$:
\begin{align*}
    r(t; T_p) =
    \begin{cases}
        0 & 0 < t \leq T_w, \\
        R \left(1 - \left(1 - \dfrac{t - T_w}{T_p}\right)^{\!3}\right) & T_w < t \leq T_w + T_p, \\
        R & T_w + T_p < t \leq T,
    \end{cases}
\end{align*}
where $R$ is the target sparsity (fraction of neurons removed), and $T_w, T_p, T$ are the number of pruning warmup steps, iterative pruning steps, and total training steps, respectively. The schedule starts at $r = 0$ (full model retained), smoothly increases to $r = R$ (target sparsity), and stays at $R$ during recovery. We set $T_w = T_l$ for integrated enlarge-and-prune pipelines, and equal to the learning rate warmup steps for naive enlarge-and-prune pipelines.

\section{Minitron: Activation-Based Pruning}
\label{appendix:minitron_importance}
We denote the activation of the $n$-th data sample of an FFN layer as
\begin{align*}
    \bZ^{(n)} = \sigma (\bX^{(n)}\bW_{\rm up}^{\top}) \odot (\bX^{(n)}\bW_{\rm gate}^{\top}),
\end{align*}
where $\bZ^{(n)} \in \RR^{l \times h}$, $l$ is the sequence length. The importance score of the $i$-th neuron is defined as
\begin{align*}
    \bc_i = \frac{1}{B}\sum_{n=1}^B \norm{\bZ_{[:, i]}^{(n)}}_2,
\end{align*}
where $B$ is the size of the calibration dataset (a random subset of the pre-training corpus), $\bZ_{[:, i]}^{(n)}$ is the $i$-th column of $\bZ^{(n)}$, and $||\cdot||_2$ is the L2-norm of a vector. Following \citet{muralidharan2024compact}, we set $B=1024$. Finally, we apply \eqref{eq:mask} with $T_p = 1$ to generate the pruning mask.

\section{Structured Pruning of Attention Heads}
\label{appendix:attn_pruning}
We describe structured pruning of attention heads in parallel with Section~\ref{sec:structured_pruning} for FFN width.
We consider a multi-head attention layer with Grouped Query Attention (GQA) \citep{ainslie2023gqa}.
Given an input $\bX \in \RR^{l \times d}$, the layer computes $n_q$ query heads and $n_{kv}$ key-value heads:
\begin{align*}
    \bZ_j &= \text{Attention}(\bX \bW_{Q_j}^{\top},\; \bX \bW_{K_{g(j)}}^{\top},\; \bX \bW_{V_{g(j)}}^{\top}), \quad j = 1, \ldots, n_q, \\
    \text{MHA}(\bX) &= [\bZ_1; \ldots; \bZ_{n_q}] \bW_O,
\end{align*}
where $\bW_{Q_j} \in \RR^{d_h \times d}$ is the query projection for the $j$-th head, $\bW_{K_{g(j)}}, \bW_{V_{g(j)}} \in \RR^{d_h \times d}$ are the shared key and value projections for KV group $g(j)$, $d_h$ is the per-head dimension, $\bW_O \in \RR^{d \times n_q d_h}$ is the output projection, and $g: \{1,\ldots,n_q\} \to \{1,\ldots,n_{kv}\}$ maps each query head to its KV group.

\textbf{Head importance scoring.}
Following the sensitivity-based approach in Section~\ref{method:iterative_pruning}, we compute the importance of each query head $j$ by aggregating element-wise sensitivity scores over its query projection matrix:
\begin{align}
\label{eq:head_impt}
    \bc_j^{(t)} = f_1\!\left(\tilde{\bS}_{Q_j}^{(t)}\right),
\end{align}
where $\tilde{\bS}_{Q_j}^{(t)} \in \RR^{d_h \times d}$ is the moving-average sensitivity score matrix for $\bW_{Q_j}$ (computed as in \eqref{eq:moving_avg}), and $f_1: \RR^{d_h \times d} \rightarrow \RR$ aggregates element-wise scores to a scalar (e.g., $\max$ or $\text{mean}$).

\textbf{Pruning mask and weight compression.}
We define a binary mask $\bbm_{\rm attn} \in \{0,1\}^{n_q}$ over query heads using the same sparsity schedule as \eqref{eq:mask}:
\begin{align*}
  \bbm_{{\rm attn},j}^{(t)} =
  \begin{cases}
    1 & \text{if $\bc_{j}^{(t)}$ is in top $(1 - r(t; T_p))$ fraction of $\bc^{(t)}$}, \\
    0 & \text{otherwise}.
  \end{cases}
\end{align*}
A KV group $k$ is retained if at least one of its associated query heads is retained:
\begin{align*}
    \bbm_{{\rm kv},k}^{(t)} = \max_{j:\, g(j)=k} \bbm_{{\rm attn},j}^{(t)}.
\end{align*}
The pruned weight matrices are obtained by selecting the rows/columns corresponding to retained heads:
\begin{align*}
    \bW_{Q_j}^{'} &= \bW_{Q_j} \cdot \bbm_{{\rm attn},j}^{(t)}, \quad j = 1,\ldots,n_q, \\
    \bW_{K_k}^{'} &= \bW_{K_k} \cdot \bbm_{{\rm kv},k}^{(t)},\quad
    \bW_{V_k}^{'} = \bW_{V_k} \cdot \bbm_{{\rm kv},k}^{(t)}, \quad k = 1,\ldots,n_{kv}, \\
    \bW_O^{'} &= \bW_O[\,:\,,\; \{j \cdot d_h, \ldots, (j{+}1)d_h - 1\}_{j:\bbm_{{\rm attn},j}=1}].
\end{align*}
After removing masked-out rows and columns, the pruned layer has $n_q^{'}$ query heads and $n_{kv}^{'} \leq n_{kv}$ KV groups, where $n_q^{'} = \|\bbm_{\rm attn}\|_1$.
The attention hidden dimension reduces from $n_q d_h$ to $n_q^{'} d_h$ while keeping the per-head dimension $d_h$ fixed, preserving the model's positional encoding and layer structure.

In our experiments (Section~\ref{sec:prune_attn}), we enlarge the number of query and KV heads by a factor of $2\times$ or $3\times$ and prune back to the target head count using iterative importance-based pruning as described above.

\section{Detailed Experiment Setups}
\label{appendix:experiment_setups}

\subsection{Model Architectures}
\label{appendix:model_arch}                                                           
  We present the key configurations of the models in Table~\ref{tab:model_arch}.        
                                                                                        
\begin{table}[htb!]
\begin{small}
\begin{center}
\vspace{-4mm}
\caption{\label{tab:model_arch}Model configurations. \# Params is non-embedding
parameter count. Per-head dimension is 64 for all models. For GQA, \# KV Heads gives
the shared key/value group count. Attention hidden size equals \# Q Heads $\times$
Head Dim.}
\setlength{\tabcolsep}{4pt}
\begin{tabular}{l|ccccccccc}
\toprule                                                                       
\multirow{2}{*}{\bf Model} & \multirow{2}{*}{\bf \# Params} & {\bf Input}  & {\bf Model} & {\bf \# Q}    & {\bf \# KV}   & {\bf Head} & {\bf Attn}   & {\bf FFN} & \multirow{2}{*}{\bf \# Layers} \\  
~   & ~ & {\bf Length} & {\bf Dim} & {\bf Heads}   & {\bf Heads}   & {\bf Dim}  & {\bf Hidden} & {\bf Hidden} & ~ \\  
\midrule
300m         & 302M  & 4096 & 1024 & 16 & 8  & 64 & 1024 & 3072  & 24 \\
300m-attn2x  & 378M  & 4096 & 1024 & 32 & 16 & 64 & 2048 & 3072  & 24 \\
300m-attn3x  & 453M  & 4096 & 1024 & 48 & 24 & 64 & 3072 & 3072  & 24 \\
300m-ffn8x       & 679M  & 4096 & 1024 & 16 & 8  & 64 & 1024 & 8192  & 24 \\
300m-f8x-a2x     & 755M  & 4096 & 1024 & 32 & 16 & 64 & 2048 & 8192  & 24 \\
\midrule                                                                       700m         & 717M  & 4096 & 1536 & 24 & 8  & 64 & 1536 & 5120  & 24 \\
700m-ffn8x       & 1.51B & 4096 & 1536 & 24 & 8  & 64 & 1536 & 12288 & 24 \\
\midrule
1b           & 1.21B & 4096 & 2048 & 32 & 8  & 64 & 2048 & 6528  & 24 \\
1b-ffn8x         & 2.67B & 4096 & 2048 & 32 & 8  & 64 & 2048 & 16384 & 24 \\
\midrule
2b           & 1.98B & 4096 & 2560 & 40 & 8  & 64 & 2560 & 8704  & 24 \\
2b-ffn8x         & 4.15B & 4096 & 2560 & 40 & 8  & 64 & 2560 & 20480 & 24 \\
\midrule
3b           & 2.91B & 4096 & 3072 & 48 & 8  & 64 & 3072 & 10752 & 24 \\
3b-ffn8x         & 5.96B & 4096 & 3072 & 48 & 8  & 64 & 3072 & 24576 & 24 \\
\bottomrule
\end{tabular}
\end{center}
\vspace{-3mm}
\end{small}
\end{table}

\subsection{Datasets}
We train all models on DCLM \citep{li2024datacomp}, an open-source pretraining corpus containing 4T unique tokens of diverse domains.

\subsection{Baselines}
\label{appendix:baselines}
We compare against two categories of baselines:

\noindent {\bf (i) Training from scratch.} We train the target-size model (1.3B) from scratch for 1T and 2T tokens.

\noindent {\bf (ii) Naive enlarge-and-prune pipelines.} The 2.8B enlarged model is pretrained for 1T tokens, pruned to 1.3B using one of the following techniques, and then continually pretrained for another 1T tokens.

$\bullet$ {\it One-shot random pruning (OSRP)} randomly generates a pruning mask satisfying the target sparsity in a single step.

$\bullet$ {\it Minitron} computes the pruning mask based on FFN activations in a single pass (see Appendix~\ref{appendix:minitron_importance}). We do not apply the distillation introduced in its original method.

$\bullet$ {\it Sheared LLaMA} \citep{xia2023sheared} learns the pruning mask via auxiliary parameters and modified training objectives. The pruning process consumes 100B tokens. We do not apply the dynamic batch loading introduced in its original method.

\subsection{Training Hyperparameters}
\label{appendix:hyperparams}
All experiments use the AdamW optimizer \citep{loshchilov2017decoupled} with a sequence length of 4096 and a batch size of 1024, regardless of model size or pruning method.
Peak and end learning rates follow optimal hyperparameter transfer \citep{yang2022tensor}.
Table~\ref{tab:hyperparams} lists the learning rate schedule and weight decay for each configuration.

\begin{table}[htb]
\centering
\vspace{-5mm}
\caption{\label{tab:hyperparams} Training hyperparameters. All configurations use AdamW with a cosine schedule and linear warmup.}
\begin{small}
\setlength{\tabcolsep}{4pt}
\begin{tabular}{l|cccc}
\toprule
{\bf Configuration} & {\bf Peak LR} & {\bf End LR} & {\bf Warmup Steps} & {\bf Weight Decay} \\
\midrule
300m from scratch          & 7.5e-3  & 3.75e-5 & 5000 & 3.16e-4 \\
300m enlarged (302m8x)     & 7.5e-3  & 3.75e-5 & 5000 & 3.16e-4 \\
302m-IDEA Prune (302m8x)   & 7.5e-3  & 3.75e-5 & 5000 & 3.16e-4 \\
\midrule
700m from scratch          & 5.0e-3  & 2.5e-5  & 5000 & 3.16e-4 \\
700m enlarged (700m8x)     & 5.0e-3  & 2.5e-5  & 5000 & 3.16e-4 \\
700m-IDEA Prune (700m8x)   & 5.0e-3  & 2.5e-5  & 5000 & 3.16e-4 \\
\midrule
1b from scratch            & 3.75e-3 & 1.88e-5 & 5000 & 3.16e-4 \\
1b enlarged (1b8x)         & 3.75e-3 & 1.88e-5 & 5000 & 3.16e-4 \\
1b-IDEA Prune (1b8x)       & 3.75e-3 & 1.88e-5 & 5000 & 3.16e-4 \\
\midrule
2b from scratch            & 3.0e-3  & 1.5e-5  & 5000 & 3.16e-4 \\
2b enlarged (2b8x)         & 3.0e-3  & 1.5e-5  & 5000 & 3.16e-4 \\
2b-IDEA Prune (2b8x)       & 3.0e-3  & 1.5e-5  & 5000 & 3.16e-4 \\
\midrule
3b from scratch            & 2.5e-3  & 1.25e-5 & 5000 & 3.16e-4 \\
3b enlarged (3b8x)         & 2.5e-3  & 1.25e-5 & 5000 & 3.16e-4 \\
3b-IDEA Prune (3b8x)       & 2.5e-3  & 1.25e-5 & 5000 & 3.16e-4 \\
\bottomrule
\end{tabular}
\vspace{-4mm}
\end{small}
\end{table}

\subsection{Hardware and Compute}
\label{appendix:compute}
All experiments are conducted on Google Cloud v5p TPU accelerators using the AXLearn framework \citep{axlearn2024}.
Training the 2.8B model for 1T tokens consumes approximately $2.5 \times 10^{22}$ FLOPs.
The total compute for the main {\our} experiment (2T tokens end-to-end) is approximately $3.8 \times 10^{22}$ FLOPs, comparable to the naive enlarge-and-prune baselines which consume $2.5 \times 10^{22}$ (enlarged pretraining) + $1.3 \times 10^{22}$ (recovery) FLOPs.

{\bf Wall-clock time.}
On TPU v5p-512 with the batch size fixed at 1024 across settings, the step time is 0.60s for training the 1.3B model, 0.58s for training the 2.8B model, and 0.60s for pruning 2.8B to 1.3B, each with a 2T-token budget.
At this scale, a training step is bottlenecked by fixed memory and communication overhead rather than compute, so the wider FFN of the 2.8B model fills otherwise idle matrix-unit cycles (MFU roughly doubles) and its extra FLOPs are nearly free.
Hence, tokens are a fair cost metric in our setting.
We note that MFU is not matched across settings. We therefore report tokens and FLOPs as the primary cost metrics, following \citet{muralidharan2024compact, xia2023sheared, li2020train}.
Since pruning removes whole neurons, the compacted weight matrices are ordinary dense matrices of smaller width, requiring no masking or gather/scatter operations.

\section{Additional Hyperparameter Ablations}
\label{appendix:hyperparam_ablation}

\begin{figure}[htb!]
\centering
\begin{subfigure}{0.47\linewidth}
    \centering
    \includegraphics[width=1.0\linewidth]{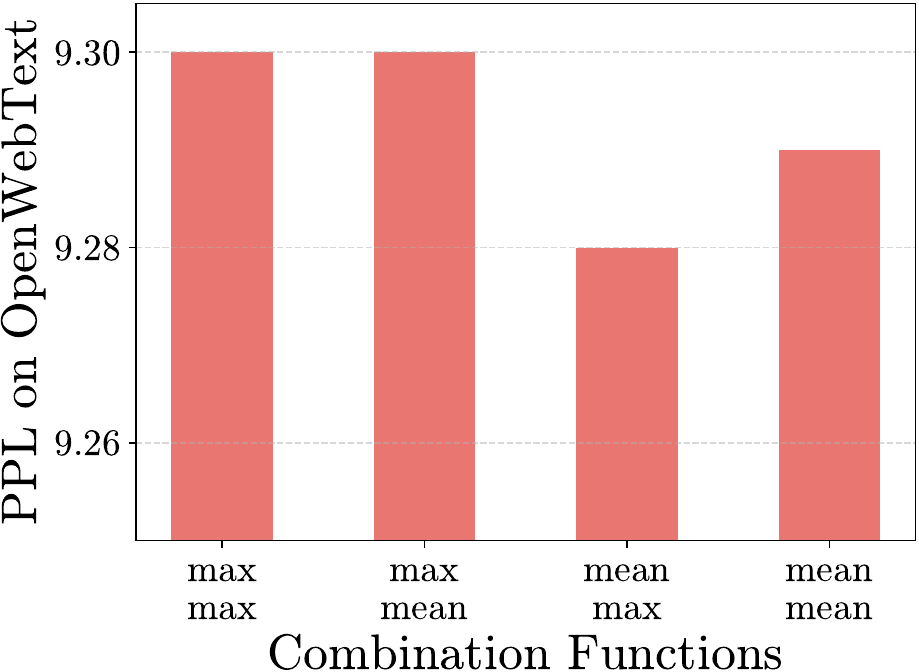}
    \caption{\label{fig:impt_score_comb} Importance Combination}
\end{subfigure}
\hfill
\begin{subfigure}{0.47\linewidth}
    \centering
    \includegraphics[width=1.0\linewidth]{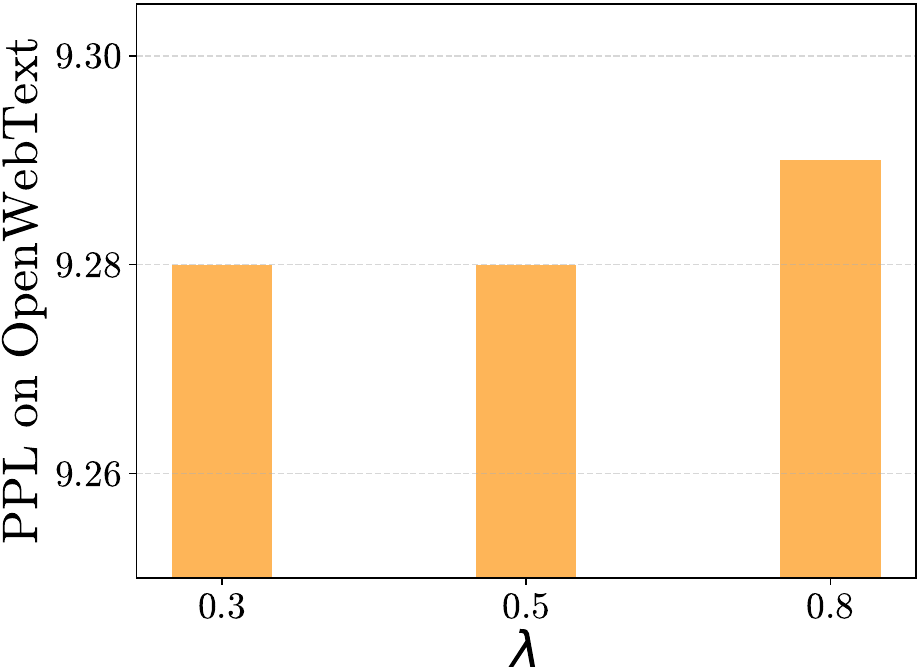}
    \caption{\label{fig:lambda} EMA Coefficient $\lambda$}
\end{subfigure}
\vspace{-3mm}
\caption{\label{fig:hyperparameter}Additional hyperparameter ablations. ``Mean-max'' is the best choice of $f_1$ and $f_2$. The EMA coefficient $\lambda$ is robust from 0.3 to 0.8.}
\end{figure}

{\bf Importance Score Combination.} We study how the choice of importance score combination function $f_1(\cdot)$ and $f_2(\cdot)$ in \eqref{eq:neuron_impt} affects model performance. We try $\max(\cdot)$ and ${\rm mean}(\cdot)$ for each of $f_1(\cdot)$ and $f_2(\cdot)$. As shown in Figure~\ref{fig:impt_score_comb}, ``mean-max'' is the best choice, though the others do not show significant degradation.

{\bf Moving Average Coefficient.} We investigate the impact of the coefficient $\lambda$ in \eqref{eq:moving_avg}, exploring values of 0.3, 0.5, and 0.8. As shown in Figure~\ref{fig:lambda}, the optimal $\lambda$ lies between 0.3 and 0.5, with 0.8 also acceptable.

\section{Results of Final Stage}
\label{appendix:final_stage}
To assess convergence stability, we evaluate the 1.3B model trained from scratch (1T tokens) at the final three checkpoints---corresponding to 0.96T, 0.98T, and 1.00T training tokens.
As shown in Table~\ref{tab:last_3_ckpt_results}, the results are remarkably consistent across checkpoints, with variations typically within 0.2 points.
This confirms that the training has entered the convergence regime and that performance noise at this stage is minimal.
We use these scratch-model results as a proxy to demonstrate convergence stability in the 2.8B$\to$1.3B setting; analogous final-stage results for {\our} are not available.

\begin{table}[tbh!]
\centering
\vspace{-4mm}
\caption{Convergence stability of the 1.3B model trained from scratch for 1T tokens, evaluated at the final three checkpoints.}
\label{tab:last_3_ckpt_results}
\begin{tabular}{c|ccccc}
\toprule
\textbf{Eval at (tokens)} & \textbf{OWT $\downarrow$} & \textbf{Arc-C $\uparrow$} & \textbf{HellaSwag $\uparrow$} & \textbf{TriviaQA $\uparrow$} & \textbf{MMLU$\uparrow$} \\
\midrule
0.96T   & 9.10 & 39.59 & 52.84 & 30.03 & 28.72 \\
0.98T   & 9.10 & 38.40 & 52.69 & 30.56 & 29.12 \\
1.00T   & 9.10 & 39.76 & 52.81 & 30.33 & 28.95 \\
\midrule
Average & 9.10 & 39.25 & 52.78 & 30.31 & 28.93 \\
\bottomrule
\end{tabular}
\end{table}

\section{Additional Benchmarks}
\label{appendix:more_benchmarks}
We further evaluate the 1.3B models of Table~\ref{tab:pretraining-pruning} on Arc-Easy \citep{allenai2018arc}, WinoGrande \citep{sakaguchi2021winogrande}, and CommonsenseQA (CSQA) \citep{talmor2019commonsenseqa}.
As shown in Table~\ref{tab:more_benchmarks}, {\our} is better than or on par with every baseline.
Similar to MMLU in Table~\ref{tab:pretraining-pruning}, naive pipelines fall far behind on CSQA, while {\our} matches the scratch baseline.

\begin{table}[htb!]
\centering
\vspace{-4mm}
\caption{\label{tab:more_benchmarks} Accuracy($\uparrow$) of 1.3B models on additional benchmarks. All models use a 2T-token budget.}
\begin{small}
\begin{tabular}{c|ccc}
\toprule
{\bf Method} & {\bf Arc-Easy} & {\bf WinoGrande} & {\bf CSQA} \\
\midrule
{1.3B-2T from scratch} & {\bf 73.8} & {66.7} & {50.04} \\
\midrule
{OSRP-2T (1.3B)} & {72.8} & {67.0} & {25.88} \\
{Minitron-2T (1.3B)} & {73.1} & {\bf 67.4} & {26.70} \\
{Sheared LLaMA-2T (1.3B)} & {73.4} & {66.8} & {27.02} \\
{\our}-2T (1.3B) & {73.2} & {\bf 67.4} & {\bf 50.45} \\
\bottomrule
\end{tabular}
\end{small}
\end{table}

\end{document}